\documentclass[10pt,twocolumn,letterpaper]{article}

\usepackage[pagenumbers]{cvpr}

\usepackage{mathtools}
\usepackage{colortbl}
\usepackage{tabularx}

\newcommand{\ours}{T4CT\xspace}
\newcommand{\PI}{\ensuremath{\pi_{0.5}}\xspace}

\newcommand{\stopgrad}{\operatorname{sg}}
\newcommand{\E}{\mathbb{E}}
\newcommand{\R}{\mathbb{R}}
\newcommand{\calN}{\mathcal{N}}
\newcommand{\appref}[1]{\cref{#1}}
\DeclareMathOperator*{\argmin}{arg\,min}

\newcommand{\papertablefont}{\fontsize{8.5}{10.2}\selectfont\renewcommand{\arraystretch}{1.08}\setlength{\heavyrulewidth}{0.7pt}\setlength{\lightrulewidth}{0.4pt}}

\definecolor{tabours}{HTML}{ECEFF8}
\definecolor{tabgroup}{HTML}{5A606B}

\usepackage{iftex}
\ifPDFTeX\else
    \usepackage{fontspec}
\fi

\makeatletter
\g@addto@macro\normalsize{%
  \setlength{\abovedisplayskip}{6pt plus 2pt minus 2pt}%
  \setlength{\belowdisplayskip}{6pt plus 2pt minus 2pt}%
  \setlength{\abovedisplayshortskip}{3pt plus 2pt}%
  \setlength{\belowdisplayshortskip}{4pt plus 2pt minus 2pt}}
\makeatother
\normalsize
\renewcommand{\papertablefont}{\fontsize{8.5}{10.2}\selectfont\renewcommand{\arraystretch}{1.0}\setlength{\heavyrulewidth}{0.7pt}\setlength{\lightrulewidth}{0.4pt}}

\definecolor{cvprblue}{rgb}{0.21,0.49,0.74}
\usepackage[pagebackref,breaklinks,colorlinks,allcolors=cvprblue,hypertexnames=false]{hyperref}

\title{T4CT: Learning to Act from Test-Time Associative Memory}

\author{%
Youngju Na \quad Sangmin Lee \quad Taegeun Yang \quad Taeyeon Kim\\
Sebin Lee \quad Jumin Lee \quad Yoonki Cho \quad Sung-Eui Yoon\textsuperscript{$\dagger$}\\[2pt]
KAIST\\[2pt]
{\small\url{https://youngju-na.github.io/T4CT/}}
}

\begin{document}

\twocolumn[{%
\renewcommand\twocolumn[1][]{#1}%
\maketitle
\begin{center}
    \includegraphics[width=\textwidth]{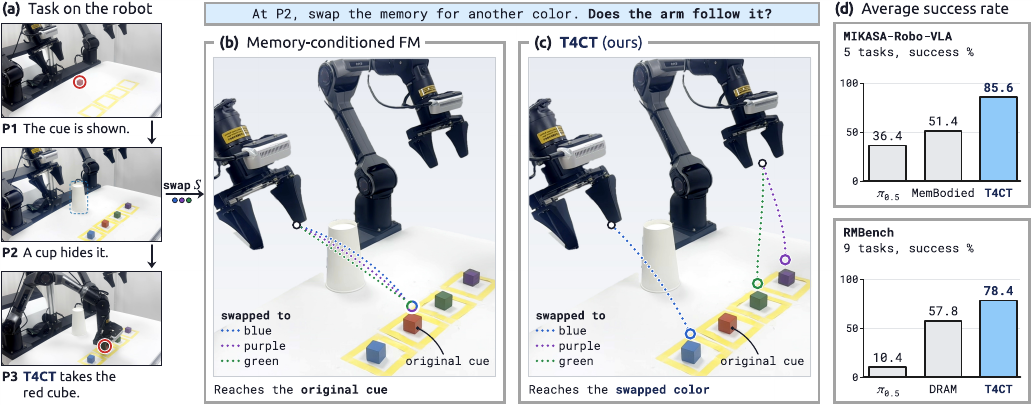}
    \captionof{figure}{
    \textbf{(a)} On a bimanual robot, (P1) a red cue cube is shown, (P2) it is hidden under a cup, and (P3) the arm grasps the matching cube. At P2, we replace the policy's memory $S$ with one stored from a cue of each of the other three colors.
    \textbf{(b)} With \emph{memory-conditioned FM}, trained with the flow-matching (FM) loss alone, the arm still reaches toward the original cue across swaps and follows the new memory in only 2 of 9 runs.
    \textbf{(c)} With \ours, the arm reaches the cube matching the color in memory in all 9 runs (3 memory swaps $\times$ 3 noise draws).
    \textbf{(d)} Average success rate on MIKASA-Robo-VLA~\cite{cherepanov2026mikasarobovla} and RMBench~\cite{chen2026rmbench}.}
    \label{fig:teaser}
\end{center}

\vspace{6pt}%
}]
\begingroup\renewcommand{\thefootnote}{}\footnotetext{\textsuperscript{$\dagger$}Correspondence: \texttt{sungeui@kaist.ac.kr}}\endgroup
 \begin{abstract}
Vision-language-action policies can retain past cues yet fail to use them when selecting actions.
Our analysis identifies one source of this gap: partially noised training actions can reveal the intended choice during flow-matching training, reducing the incentive to learn how action selection depends on memory.
We propose \ours, which trains test-time associative memory through an action readout and uses this readout to initialize flow generation.
An action readout predicts a chunk from the current observation and retrieved memory without access to the noised training action.
Our analysis demonstrates that direct readout supervision drives cue-guided action selection, while initializing the flow from the readout pairs that choice with continuous refinement to improve over starting from a Gaussian source.
Across MIKASA-Robo-VLA and RMBench, \ours achieves competitive or higher average success than prior memory-augmented baselines, and we further evaluate it on real-world non-Markovian manipulation tasks.
\end{abstract}

\section{Introduction}
\label{sec:intro}

Vision-language-action (VLA) policies typically generate actions from the current observation and instruction~\cite{black2024pi0,physicalintelligence2025pi05}.
In non-Markovian manipulation tasks, however, earlier cues often disappear over time, leaving the current observation compatible with several distinct actions (action modes) and requiring the policy to rely on episode history.
To retain information across an episode, memory-augmented policies often maintain banks of past frames or tokens~\cite{shi2025memoryvla,sridhar2025memer}, recurrent states~\cite{cherepanov2026muvla}, or test-time associative memory via test-time training (TTT), which compresses observed context into fixed-size fast weights updated during inference~\cite{sun2024tttlayers,jiang2026robottt,liu2026t2mem,pala2026membodied,zhao2026dram}.

In these associative-memory policies, an online inner update continually writes episode history into the fast weights and reads from them at each step, while an outer-loop action objective trains the memory projections to learn which historical cues to store and retrieve.
Typically, the retrieved memory conditions a flow-matching action expert that transforms Gaussian noise into an action chunk~\cite{lipman2023flowmatching}.
Yet this memory-conditioned policy does not necessarily rely on stored cues when selecting among action modes.
For example, in the color-recall task of \cref{fig:teaser}, a color cue is covered by a cup and the robot must pick the matching cube among several colored cubes.
Even when the fast weights retain the earlier cue, the policy often reaches for the wrong cube, and replacing the stored memory with another color rarely changes which cube it selects.

To understand this behavior, we examine how flow-matching training can lead policies to ignore stored memory. During training, the action expert predicts a velocity from a partially noised action $x_\tau$ (\ie, an action-noise interpolant) formed by mixing the demonstrated chunk with Gaussian noise.
Except at high noise levels, $x_\tau$ already reveals the demonstrated action mode, allowing the policy to reduce its velocity loss with little reliance on memory, a route we call the \emph{interpolant shortcut}.
Moreover, once the arm begins moving toward a target, the current observation itself shows which target is being approached, so predicting the remaining motion no longer requires memory.
As a result, memory helps reduce the training loss only in a narrow decision window, at high noise and in the first few frames of a movement.
Outside this narrow window, the flow-matching loss provides little incentive to use memory, diluting the gradients needed to learn how retrieved cues should guide action selection at deployment.

To address this challenge, we introduce \ours (\emph{test-time training to act}), a framework for learning to act from test-time associative memory by decoupling history-dependent mode selection from flow-based trajectory refinement.
Specifically, a lightweight readout decoder predicts an action chunk, the \emph{action readout}, directly from the current observation and retrieved memory without access to the noised demonstration.
Without the interpolant shortcut, its loss backpropagates directly through memory reads and updates, supervising the associative memory to store and retrieve the cues needed for action selection.
To recover fine-grained continuous control, we center the flow source distribution on the action readout and train the action expert to refine samples from this source into demonstrated actions.

Across MIKASA-Robo-VLA~\cite{cherepanov2026mikasarobovla}, RMBench~\cite{chen2026rmbench}, and real-world dual-arm tasks with occluded cues, \ours achieves competitive or higher mean success than prior memory-augmented baselines.
Diagnostic interventions and ablations further indicate that generated actions follow retrieved cues rather than shortcut correlations, with both readout supervision and flow initialization contributing to this behavior.
Our contributions are as follows:
\begin{itemize}[leftmargin=*,itemsep=1pt,topsep=2pt]
    \item We identify the \emph{interpolant shortcut}, by which the action-noise interpolant reveals the intended choice during flow-matching training, and show that the resulting memory gain is confined to a narrow decision window.
    \item We propose \ours, which uses an action readout to train test-time associative memory and initialize flow generation, restoring credit assignment for history-dependent mode selection while preserving expressive trajectory refinement.
    \item We evaluate \ours across two simulation benchmarks and real-world dual-arm manipulation tasks, using counterfactual memory interventions and component ablations to verify that generated actions causally follow stored cues.
\end{itemize}

\section{Related Work}
\label{sec:related}

\vspace{-3pt}\paragraph{Explicit memory for robot policies.}
For non-Markovian tasks that require episodic context, recent robot policies with explicit memory retain a buffer of past tokens or frames alongside the current view to generate history-conditioned actions.
For example, MemoryVLA~\cite{shi2025memoryvla} maintains a bank of perceptual and cognitive tokens, SAM2Act~\cite{fang2025sam2act} and HAMLET~\cite{koo2026hamlet} store spatial keyframes or per-step moment tokens, and EventVLA~\cite{yang2026eventvla} records event-driven visual keyframes across long horizons.
Storing past observations preserves visual detail, but attending over a growing buffer increases inference cost with episode length.
Heuristic consolidation~\cite{shi2025memoryvla} and VLM-based frame selection~\cite{sridhar2025memer,mark2026bpp} can bound this cost, yet their selection criteria may not generalize across tasks and can discard cues needed for later decisions.

\vspace{-3pt}\paragraph{Recurrent and test-time associative memory.}
Recurrent policies compress history into fixed-size memory tokens~\cite{cherepanov2026rate,cherepanov2026muvla} or state-space representations~\cite{zhou2025mtil}.
Linear recurrent models use input-dependent updates to control how past information is retained and new information is incorporated~\cite{gu2023mamba,yang2024gatedlinear,yang2025gateddelta}.
Test-time training (TTT) extends this recurrent paradigm by compressing observed context into the parameters of a queryable model, known as fast weights~\cite{sun2024tttlayers,behrouz2025titans,zhang2025lact,jin2026zipmap}.
An inner objective fits a mapping from keys to values, storing their associations in the fast weights (write).
Evaluating the updated model on queries retrieves information from this compressed context (read)~\cite{wang2025testtimeregression,behrouz2025miras,jin2026zipmap}.
Recent robot policies use such associative memories to accumulate episode history and condition action generation on the retrieved information~\cite{jiang2026robottt,feng2026wamttt,zhao2026dram,pala2026membodied}.
During offline training, an outer-loop action objective trains the write and retrieval representations through memory updates, yet when paired with flow matching, it does not guarantee that retrieved cues govern action selection.

\vspace{-3pt}\paragraph{Shortcut learning and condition neglect.}
In imitation learning, policies can lower training loss via spurious correlations while underusing conditioning inputs.
Such behavior is related to shortcut learning~\cite{geirhos2020shortcut} and causal confusion~\cite{dehaan2019causal}.
History-conditioned policies, for instance, may exploit temporal correlations by copying previous actions~\cite{wen2020fighting}.
To mitigate condition neglect, prior work adds auxiliary objectives such as past-action prediction~\cite{torne2025ptp} or builds a condition-dependent flow source~\cite{dong2025conditioning}.
However, these approaches do not explain why flow-matching supervision~\cite{lipman2023flowmatching,black2024pi0,physicalintelligence2025pi05} can leave episodic memory underused.
In this work, we analyze how the action-noise interpolant in flow-matching policies~\cite{lipman2023flowmatching,black2024pi0,physicalintelligence2025pi05} weakens the learning signal for memory-dependent mode selection.
To address this, we separate memory-guided action prediction from flow-based trajectory refinement.

\section{Motivation}
\label{sec:analysis}
In this section, we formalize test-time memory in flow policies (\cref{sec:prelim}) and analyze why training in standard flow matching may provide limited incentive to use remembered information for action selection (\cref{sec:memory_unused}).

\subsection{Problem Formulation}
\label{sec:prelim}

\vspace{-3pt}\paragraph{Setup.}
We investigate vision-language-action (VLA) policies that generate an $H$-step chunk of $d_a$-dimensional actions from the current observation $o_t$ (images, robot state, and instruction) and a memory state $m_t$ summarizing past observations $o_{1:t-1}$,
\begin{equation}
a_t \sim \pi_\theta(a_t \mid o_t, m_t), \qquad a_t \in \mathbb{R}^{H \times d_a}.
\label{eq:policy}
\end{equation}
Memory provides historical context for resolving \emph{perceptual aliasing}~\cite{whitehead1991learning}, where similar current observations $o_t$ correspond to distinct task states requiring different actions.
In this work, we obtain $m_t$ from a test-time associative memory~\cite{sun2024tttlayers,wang2025testtimeregression}, which compresses past observations into episode-specific fast weights online and relies on outer-loop action supervision during training to learn which historical cues to store and retrieve.

\vspace{-3pt}\paragraph{Flow matching.}
Flow-matching policies realize $\pi_\theta$ by transporting Gaussian noise into an action chunk along a learned velocity field $v_\theta$, where the outer-loop objective takes the form of velocity prediction~\cite{lipman2023flowmatching,black2024pi0,physicalintelligence2025pi05}.
Specifically, training mixes a demonstrated action chunk $a_t^\star$ with Gaussian noise at flow time $\tau\in[0,1]$:
\begin{equation}
x_\tau = (1-\tau) a_t^\star + \tau \epsilon, \qquad \epsilon \sim \mathcal{N}(0, I).
\label{eq:interpolant}
\end{equation}
At $\tau=0$, the interpolant $x_\tau$ (\cref{eq:interpolant}) is the demonstrated action itself, while at $\tau=1$ it is pure Gaussian noise.
The velocity field $v_\theta$ learns to predict the target velocity $\epsilon - a_t^\star$ from $x_\tau$ and context $(o_t, m_t)$:
\begin{equation}
\mathcal{L}_{\mathrm{FM}} = \mathbb{E}\left[ \left\| v_\theta(x_\tau, \tau; o_t, m_t) - (\epsilon - a_t^\star) \right\|_F^2 \right].
\label{eq:fm_loss}
\end{equation}
During training, we randomly sample $\tau$ and supervise $v_\theta$ on $x_\tau$.
At inference, the policy samples $a_t \sim \pi_\theta$ by integrating $v_\theta$ from Gaussian noise $x_1\sim\mathcal N(0,I)$ at $\tau=1$ to $\tau=0$.

\vspace{-3pt}\paragraph{Memory as condition.}
\label{sec:memcond}
A common design provides the memory read $m_t$ as an additional conditioning prefix alongside images, robot state, and language, while the action tokens are generated as a suffix~\cite{shi2025memoryvla}.
For example, memory reads can be represented as tokens that the action expert attends to together with the observation tokens.
Generation still starts from Gaussian noise, and training uses the same flow-matching objective (\cref{eq:fm_loss}).
We refer to this baseline design as \emph{memory-conditioned FM}.

\begin{figure}[t]
    \centering
    \includegraphics[width=\linewidth]{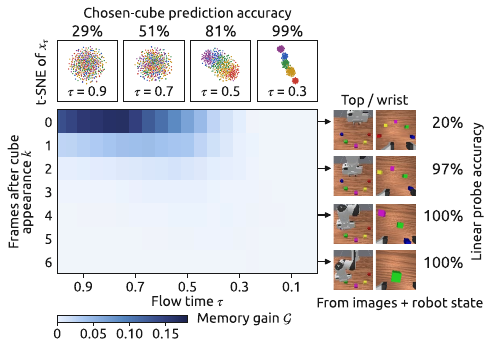}
    \caption{\textbf{Memory gain across flow time and frames.}
    The map shows the estimated memory gain $\mathcal G$ (\cref{eq:decision_gain}) per action coordinate on RememberColor5, the predictive value that memory adds beyond the interpolant $x_\tau$ and the current observation.
    The chosen cube becomes predictable from $x_\tau$ alone as noise decreases (top, points colored by target cube) and from the current observation alone within two frames.
    The gain therefore remains mainly within a narrow decision window, at high noise (large $\tau$) and in the first frames.}
    \label{fig:shortcut}
\end{figure}

\subsection{When Memory Goes Unused}
\label{sec:memory_unused}
\label{sec:present}
\label{sec:hedge}

\vspace{-3pt}\paragraph{The interpolant shortcut.}
Consider the perceptually aliased task in \cref{fig:shortcut}, where an early color cue disappears and five colored cubes appear at shuffled table locations.
At the decision frame ($k=0$), the current observation $o_t$ shows where the five cubes sit, while only memory $m_t$ can specify which color was cued.
During flow-matching training, however, the velocity field $v_\theta$ also receives the interpolant $x_\tau = (1-\tau)a_t^\star + \tau\epsilon$ (\cref{eq:interpolant}), which reveals evidence about the demonstrated action $a_t^\star$ whenever $\tau < 1$.
As shown at the top of \cref{fig:shortcut}, $x_\tau$ alone separates the five target modes across most noise levels, predicting the chosen cube with 81\,\% accuracy at $\tau=0.5$ compared with 20\,\% uniform chance.
The policy can therefore minimize $\mathcal{L}_{\mathrm{FM}}$ by following the hint in $x_\tau$ rather than relying on the earlier cue in memory $m_t$, a failure mode we call the \emph{interpolant shortcut}.
Thus, while flow matching excels at generating expressive continuous trajectories, the denoising objective can prevent memory from governing history-dependent decisions.

\vspace{-3pt}\paragraph{Exponential decay of memory gain.}
To quantify memory's predictive benefit as noise decreases, we compare the minimum expected flow-matching loss with and without memory at each flow time $\tau$ (\cref{eq:fm_loss}).
Let $R_0(\tau)$ and $R_m(\tau)$ denote these Bayes-optimal velocity risks~\cite{kang2026warmprior} without and with memory $m_t$.
Their difference defines the \emph{memory gain}:
\begin{equation}
  \mathcal G(\tau)
  = R_0(\tau)-R_m(\tau)
  = \frac{1}{\tau^2}\E\|\mu_m-\mu_0\|_F^2,
  \label{eq:decision_gain}
\end{equation}
where $\mu_0=\E[a_t^\star\mid x_\tau,o_t]$ and $\mu_m=\E[a_t^\star\mid x_\tau,o_t,m_t]$ are the optimal action estimates without and with memory.
In words, memory helps only when $m_t$ shifts the expected action beyond what $x_\tau$ and $o_t$ already reveal.
To track how this gap varies with $\tau$, consider two equally likely target chunks $a^+,a^-$ (\eg, reaching for two cubes) under the same observation $o_t$, so that only memory tells them apart.
While memory identifies the target ($\mu_m=a_t^\star$), $\mu_0$ can infer the choice only from $x_\tau$, which forms two Gaussian clouds of width $\tau$ around $(1-\tau)a^\pm$ that separate and tighten as $\tau$ decreases (top of \cref{fig:shortcut}).
Bounding $\E\|\mu_m-\mu_0\|_F^2$ by the squared error of a nearest-mode predictor on $x_\tau$ and applying a Gaussian tail bound gives (full derivation in \appref{app:memory_gain}):
\begin{equation}
  \frac{\mathcal G(\tau)}{\mathcal G(1)}
  \leq \frac{2}{\tau^2}
  \exp\!\left[
    -\frac{(1-\tau)^2\|a^+-a^-\|_F^2}{8\tau^2}
  \right],
  \label{eq:gain_decay}
\end{equation}
where $\mathcal G(1)$ is the memory gain at pure noise.
When the two chunks are well separated (large $\|a^+-a^-\|_F$), the exponential factor overwhelms $1/\tau^2$ at moderate noise, so memory's predictive benefit does not fade gradually across noise levels but drops sharply outside the high-noise range.

\vspace{-3pt}\paragraph{Narrow decision window.}
The current observation can also reveal a history-dependent choice once the robot begins acting on that choice.
As the arm approaches a cube, its configuration indicates which cube the demonstrator selected, allowing the policy to predict the remaining movement without retrieving the earlier cue.
On the right of \cref{fig:shortcut}, a linear probe on features of current images and robot state identifies the selected cube at chance level (20\,\%) at cube appearance ($k=0$), but reaches 97\,\% accuracy just two frames later ($k=2$).
Predicting the continuation of an ongoing movement may therefore rely less on memory than deciding which movement to initiate.
Consequently, the estimated memory gain $\mathcal G$ in \cref{fig:shortcut} concentrates within a narrow decision window near the decision frame and at high noise (see \appref{app:present} for detailed analysis).
Because most training updates fall outside this window, the action expert can lower its loss from $o_t$ and $x_\tau$ alone, leaving memory conditioning prone to collapse analogous to posterior collapse~\cite{chen2017variational,dong2025conditioning}.

\begin{figure*}[t]
  \centering
  \includegraphics[width=\textwidth]{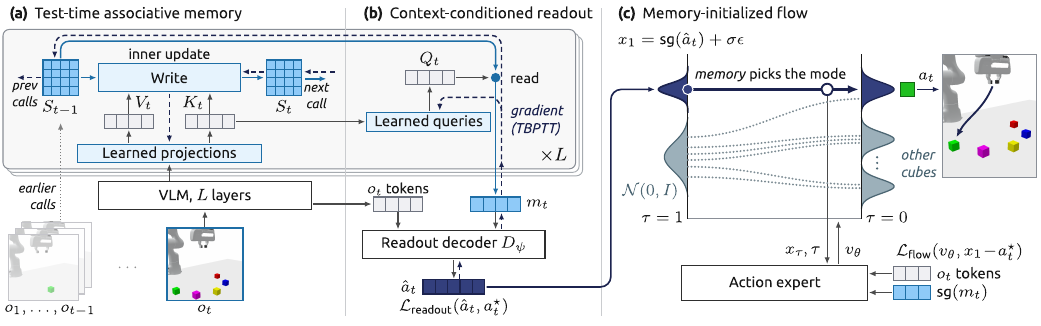}
    \caption{\textbf{Overview of \ours.}
\textbf{(a)~Test-time associative memory.} Features of the current frame $o_t$ from $L$ VLM layers are projected into keys $K_t$ and values $V_t$, which update the fast weights $S_{t-1} \to S_t$ by closed-form key--value regression.
\textbf{(b)~Context-conditioned readout.} Queries $Q_t(o_t)$ read $S_{t-1}$ to retrieve memory features $m_t$, and a decoder $D_\psi$ predicts an action chunk $\hat a_t$. The readout loss $\mathcal L_{\mathrm{readout}}$ (dashed arrows) trains both the write and the read.
\textbf{(c)~Memory-initialized flow.} The flow starts from $x_1 = \stopgrad(\hat a_t) + \sigma\epsilon$ instead of Gaussian noise, and the action expert refines the recalled choice into the action chunk $a_t$.}
  \label{fig:method}
\end{figure*}

\section{\ours\ -- Test-Time Training to Act}
\label{sec:method}

We introduce \ours, a VLA model that learns to
act with test-time associative memory. Given the current observation $o_t$ and past observations
encoded in fast weights $S_{t-1}$, the policy retrieves
historical context to predict an action chunk $a_t$. We train the policy to use this memory through action prediction without access to the target-revealing interpolant.
We first describe the memory representation (\cref{sec:method_write}), then introduce direct action-readout supervision (\cref{sec:method_ground}), and finally describe memory-initialized flow (\cref{sec:method_mif}).

\subsection{Test-Time Regression}
\label{sec:method_write}

In test-time regression~\cite{sun2024tttlayers,wang2025testtimeregression}, an associative memory stores history by fitting a regressor from keys to values online within an episode, and retrieves stored information by evaluating that regressor on queries.
Following this principle, we maintain an episode-specific fast-weight matrix $S_t \in \R^{d \times d}$ as a linear memory map at each of $L$ VLM layers (\cref{fig:method}a), reset at the start of every episode.
At policy call $t$, learned projections map the visual features of $o_t$ into keys $K_t$, values $V_t$, a diagonal token-weighting matrix $\Lambda_t$, and a retention gate $\gamma_t \in (\tfrac12, 1)$.
Omitting the layer index, we update the fast weights through regularized weighted least-squares regression~\cite{sun2024tttlayers,wang2025testtimeregression,zhao2026dram}:
\begin{equation}
  S_t = \argmin_S\
  \|K_tS-V_t\|_{\Lambda_t}^2
  + \|S-\gamma_t S_{t-1}\|_F^2.
  \label{eq:mif_inner}
\end{equation}
The first term fits new key--value associations weighted per token by $\Lambda_t$, while the second term penalizes deviations from the decayed state $\gamma_t S_{t-1}$ to balance acquisition and retention~\cite{behrouz2025miras}.
Following Longhorn’s proximal online regression formulation~\cite{liu2024longhorn,wang2025testtimeregression},
\cref{eq:mif_inner} yields the exact update $S_t=U(S_{t-1},o_t)$ through one linear solve per layer during both training and deployment.
Before writing $o_t$ into memory, the policy first reads from $S_{t-1}$ so that retrieval reflects strictly past observations $o_{1:t-1}$ (\cref{fig:method}b).
Because what must be recalled in non-Markovian manipulation depends on the current scene, learned query tokens first attend to the current visual keys $K_t$ to form scene-conditioned queries $Q_t(o_t)$, which read the memory map as $Q_t(o_t) S_{t-1}$.
The retrieved features then cross-attend to the visual tokens of $o_t$, aligning the recalled past cue with the objects currently visible in the scene to form $m_t$.

\subsection{Learning to Act via Deterministic Readout}
\label{sec:method_ground}

Outer-loop action supervision teaches memory which historical cues matter for the current decision.
To supervise both memory reads and earlier writes, we backpropagate action losses across consecutive episode calls using truncated backpropagation through time (TBPTT).
In flow matching, however, the interpolant is ideally free of target information at $\tau=1$ and partially reveals the target whenever $\tau<1$ (\cref{sec:memory_unused}).
Sampling $\tau$ more heavily near this endpoint may alleviate the leakage, but it shifts training away from the lower noise levels that the flow relies on for action refinement, which can compromise generation quality.
Supervising actions integrated from pure noise avoids the interpolant shortcut, but requires backpropagation through every integration step of the ODE solver, which is costly and can yield unstable gradients~\cite{park2025flow}.

We therefore separate memory supervision from flow-based trajectory refinement.
Specifically, we use a lightweight three-layer transformer decoder $D_\psi$.
Its learned action queries attend to the current observation tokens and retrieved memory $m_t$ to predict the demonstrated action chunk $\hat a_t = D_\psi(o_t,m_t)$, trained with:
\begin{equation}
  \mathcal L_{\mathrm{readout}}
  = \E\,\big\|D_\psi(o_t,m_t)-a_t^\star\big\|_F^2.
  \label{eq:mif_draft}
\end{equation}
Minimizing \cref{eq:mif_draft} estimates the conditional mean $\E[a_t^\star \mid o_t,m_t]$, which determines the optimal flow velocity of \cref{eq:fm_loss} at $\tau=1$.
The readout objective thus retains the memory gain $\mathcal G(1)$ on decision frames without the target-revealing interpolant.
As the decoder requires no ODE integration, TBPTT trains the memory and the readout decoder jointly across calls without backpropagation through the integration steps.

\subsection{Memory-Initialized Flow (MIF)}
\label{sec:method_mif}

While $\hat a_t$ predicts the history-dependent action mode, a deterministic point estimate may not capture fine-grained variations in contact motion and grasp timing~\cite{chi2023diffusion,black2024pi0}.
Memory-Initialized Flow (MIF) therefore centers the flow source distribution on $\hat a_t$ rather than using standard Gaussian noise $\calN(0,I)$ (\cref{fig:method}c).
Using a noise scale $\sigma > 0$ ($\sigma=0.1$ in our experiments), we define the source sample $x_1$ and training interpolant $x_\tau$ as:
\begin{equation}
\begin{aligned}
  x_1 &= \stopgrad(\hat a_t)+\sigma\epsilon,
  \qquad \epsilon\sim\calN(0,I),\\
  x_\tau &= (1-\tau)a_t^\star+\tau x_1.
\end{aligned}
\label{eq:mif_source}
\end{equation}
We train the velocity field $v_\theta$ with:
\begin{equation}
  \mathcal L_{\rm flow}
  = \E\left[\left\|v_\theta(x_\tau,\tau;\,o_t,\,\stopgrad(m_t))
  -(x_1-a_t^\star)\right\|_F^2\right].
  \label{eq:mif_flow_loss}
\end{equation}
The total training objective is $\mathcal L_{\rm total} = \mathcal L_{\rm readout} + \mathcal L_{\rm flow}$.
At inference, we sample $x_1 = \hat a_t + \sigma\epsilon$ and integrate $v_\theta$ from $\tau=1$ to $0$.
This formulation pairs readout-supervised mode selection with flow-based trajectory refinement.
Because $\mathcal L_{\mathrm{readout}}$ trains $\hat a_t$ without the interpolant shortcut, starting near $\hat a_t$ anchors mode selection.
The flow expert can then refine continuous motion around that prediction.
Detaching $\hat a_t$ and $m_t$ blocks direct flow-loss gradients into the readout and memory branches.
This prevents the flow objective from directly adjusting the source center to simplify the velocity target $x_1-a_t^\star$.

Initializing the flow at $\hat a_t$ also reduces the expected transport distance compared to a zero-mean Gaussian source.
Because $\epsilon$ is independent of both $\hat a_t$ and $a_t^\star$, the expected squared coupling cost satisfies:
\begin{equation}
  \E\|x_1-a_t^\star\|_F^2
    = \mathcal L_{\rm readout}+Hd_a\sigma^2,
  \label{eq:mif_coupling}
\end{equation}
where $Hd_a$ is the action-chunk dimension.
For a fixed $\sigma$, the conditional mean $\E[a_t^\star\mid o_t,m_t]$ minimizes this coupling cost over all source centers conditioned on $(o_t,m_t)$.
This is precisely the quantity estimated by $\mathcal L_{\rm readout}$ (\cref{sec:method_ground}).
This cost, in turn, upper-bounds the squared 2-Wasserstein distance between source and target, as well as the irreducible velocity risk of the flow~\cite{albergo2024couplings,kang2026warmprior} (\appref{app:mif_source}).
Minimizing $\mathcal L_{\rm readout}$ therefore directly lowers this bound.
Empirically, readout initialization also keeps success above a Gaussian source at every number of integration steps we evaluate, and with two steps it already exceeds the Gaussian source with ten (\cref{fig:mif}b).

\section{Experimental Results}
\label{sec:exp}
We evaluate \ours across two simulation benchmarks and three real-world bimanual manipulation tasks.
Our experiments examine (1)~closed-loop task success across diverse memory demands, (2)~whether generated actions respond to stored cues under counterfactual memory interventions, and (3)~ablation studies.
\subsection{Experimental Setup}
\label{sec:exp_setup}

\vspace{-3pt}\paragraph{Simulation benchmarks.}
We evaluate on MIKASA-Robo-VLA~\cite{cherepanov2026mikasarobovla,cherepanov2025mikasa} and RMBench~\cite{chen2026rmbench}.
On MIKASA-Robo-VLA, we train a policy across five tasks with 250 demonstrations per task and evaluate 50 episodes per task, spanning visual cue recall (\emph{RememberColor5} [RC5], \emph{RememberShapeAndColor3x3} [RSC]), spatial memory (\emph{ShellGamePush} [SGP], \emph{TakeItBack} [TIB]), and motion tracking (\emph{InterceptMedium} [IM]).
On RMBench, we train one policy per task from 50 demonstrations and evaluate 50 episodes across nine bimanual tasks: five $\mathrm{M}(1)$ tasks requiring recall of a single earlier observation and four $\mathrm{M}(n)$ tasks requiring state accumulation across interactions.

\begin{figure*}[t]
    \centering
    \includegraphics[width=\linewidth]{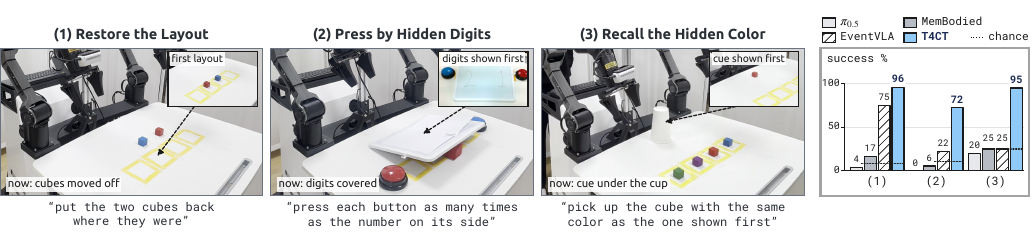}
        \caption{\textbf{Real-world tasks on a dual-arm AgileX robot.}
        (Left) \emph{Restore the Layout}, \emph{Press by Hidden Digits}, and \emph{Recall the Hidden Color}, each requiring a cue that is no longer visible when the robot acts.
        Insets show the cue presented at the start, the main views show the scene at decision time, and the quoted text is the language instruction.
        (Right) Success rate of each policy over 24, 18, and 20 trials, respectively.}
    \label{fig:curves}
\end{figure*}

\vspace{-5pt}\paragraph{Real-world setup.}
We evaluate three bimanual tasks on an AgileX Cobot Magic platform as shown in \cref{fig:curves}:
(1)~\emph{Restore the Layout}, where a red and a blue cube are placed on two of four slots (chance $1/12$) before being moved to the center, and the two arms must return each cube to its original slot;
(2)~\emph{Press by Hidden Digits}, where two digits in $\{1,2,3\}$ are briefly shown and covered (chance $1/9$), and each arm must press its button the indicated number of times; and
(3)~\emph{Recall the Hidden Color}, where a cue cube is shown and then hidden under a cup, and the corresponding arm must pick the same-colored cube among four rearranged cubes (chance $1/4$).

\vspace{-5pt}\paragraph{Baselines.}
We compare against three families of memory-augmented policies,
alongside memory-free VLAs~\cite{physicalintelligence2025pi05,kim2025openvlaoft,zheng2025xvla}.
These include explicit frame- or token-bank policies~\cite{chen2021decision,shi2025memoryvla,sridhar2025memer,yang2026eventvla},
recurrent memory models~\cite{cherepanov2026rate,cherepanov2025elmur,cherepanov2026muvla,huang2026chainvla,vanjani2026recat},
and test-time associative-memory policies~\cite{pala2026membodied,zhao2026dram}.
Published baseline results are taken from the benchmark reports~\cite{cherepanov2025mikasa,chen2026rmbench} and the original papers~\cite{cherepanov2025elmur,cherepanov2026muvla,yang2026eventvla,pala2026membodied,huang2026chainvla,vanjani2026recat,zhao2026dram}.
For controlled analysis, we also evaluate \emph{memory-conditioned FM} (\cref{sec:memcond}), which shares our \PI backbone and fast-weight memory but supplies retrieved reads as additional prefix tokens under flow matching alone.

\vspace{-5pt}\paragraph{Implementation details.}
We instantiate \ours on the pretrained \PI, adding a fast-weight memory to each of $L$ Gemma layers ($L=18$), each read by eight scene-conditioned queries, which adds about 0.5\,\% to the parameter count.
The VLM is fine-tuned with rank-16 LoRA, while the memory modules, the readout decoder, and the action expert are fully trained.
Each policy is trained on a single NVIDIA H100 or H200 GPU with about 32 policy calls per training step.
The policy takes $224\times224$ RGB frames together with the robot state and the instruction.
At inference, the flow starts from the readout and runs ten integration steps.
On MIKASA-Robo-VLA, the policy uses two cameras, predicts chunks of 8 actions, and replans at every step following $\mu$VLA~\cite{cherepanov2026muvla}.
The readout loss is backpropagated through two consecutive calls.
On RMBench, it uses three cameras, predicts chunks of 50 actions, and executes up to 50 actions per call as in prior work~\cite{chen2026rmbench,pala2026membodied}.
Following MemBodied~\cite{pala2026membodied}, the readout loss is backpropagated through windows of 7 to 18 calls.
Additional details are provided in \appref{app:mif_recipe}.

\subsection{Evaluation on Simulation Benchmarks}
\label{sec:exp_main_table}
\providecolor{tabr1}{HTML}{C8E1FF}\providecolor{tabr2}{HTML}{E1F0FF}\providecolor{tabr3}{HTML}{F2F8FF}
\begin{table}[t]
    \centering
    \caption{\textbf{MIKASA-Robo-VLA.} Success (\%) on five training tasks~\cite{cherepanov2026muvla}. Shading marks the best, second-best, and third-best values of each column.}
    \label{tab:main_mikasa}
    \papertablefont
    \fontsize{7}{8.4}\selectfont
    \setlength{\tabcolsep}{1.2pt}
    \begin{tabularx}{\linewidth}{l>{\raggedright\arraybackslash}Xlcrrrrrr}
        \toprule
        Memory & Method & VLM & Act. & RC5 & RSC & SGP & TIB & IM & Avg. \\
        \midrule
        \textcolor{tabgroup}{\textit{None}} & OpenVLA-OFT~\cite{kim2025openvlaoft} & Prism. & Reg. & 4 & 3 & 83 & 83 & 36 & 41.8 \\
        \addlinespace[2.5pt]
        \textcolor{tabgroup}{\textit{Frame bank}} & DT~\cite{chen2021decision} & -- & Reg. & 7 & 6 & 62 & 8 & \cellcolor{tabr2}56 & 27.8 \\
        \addlinespace[2.5pt]
        \textcolor{tabgroup}{\textit{Recurrent}} & RATE~\cite{cherepanov2026rate} & -- & Reg. & 13 & 8 & 78 & 42 & 32 & 34.6 \\
         & ELMUR~\cite{cherepanov2025elmur} & -- & Reg. & 19 & \cellcolor{tabr3}19 & \cellcolor{tabr3}87 & 78 & \cellcolor{tabr1}64 & \cellcolor{tabr3}53.4 \\
         & $\mu$VLA~\cite{cherepanov2026muvla} & Prism. & Reg. & \cellcolor{tabr2}93 & \cellcolor{tabr1}86 & \cellcolor{tabr2}95 & \cellcolor{tabr1}99 & 47 & \cellcolor{tabr2}84.0 \\
        \midrule
        \textcolor{tabgroup}{\textit{None}} & \PI~\cite{physicalintelligence2025pi05} & PaliG. & FM & 10 & 14 & 26 & 88 & 44 & 36.4 \\
        \addlinespace[2.5pt]
        \textcolor{tabgroup}{\textit{Frame bank}} & \PI + MemoryVLA~\cite{shi2025memoryvla} & PaliG. & FM & 19 & 6 & 74 & \cellcolor{tabr3}92 & 46 & 47.4 \\
        \addlinespace[2.5pt]
        \textcolor{tabgroup}{\textit{Recurrent}} & \PI + $\mu$VLA~\cite{cherepanov2026muvla} & PaliG. & FM & \cellcolor{tabr3}24 & 16 & 55 & 84 & 39 & 43.6 \\
        \addlinespace[2.5pt]
        \textcolor{tabgroup}{\textit{Associative}} & MemBodied~\cite{pala2026membodied} & PaliG. & FM & 23 & 13 & 79 & \cellcolor{tabr3}92 & 50 & 51.4 \\
         & \textbf{\ours (ours)} & PaliG. & FM & \cellcolor{tabr1}96 & \cellcolor{tabr2}84 & \cellcolor{tabr1}98 & \cellcolor{tabr2}98 & \cellcolor{tabr3}52 & \cellcolor{tabr1}85.6 \\
        \bottomrule
    \end{tabularx}
\end{table}

\providecolor{tabr1}{HTML}{C8E1FF}\providecolor{tabr2}{HTML}{E1F0FF}\providecolor{tabr3}{HTML}{F2F8FF}
\begin{table}[t]
    \centering
    \caption{\textbf{RMBench.} Success (\%) on nine tasks~\cite{chen2026rmbench}. M(1) tasks need one past observation and M(n) tasks several, and the averages cover each group and all nine tasks.}
    \label{tab:main_rmbench}
    \papertablefont
    \fontsize{7}{8.4}\selectfont
    \setlength{\tabcolsep}{1.5pt}
    \begin{tabularx}{\linewidth}{>{\raggedright\arraybackslash}p{2.65cm}*{9}{>{\raggedleft\arraybackslash}X}*{3}{r}}
        \toprule
        & \multicolumn{5}{c}{M(1)} & \multicolumn{4}{c}{M(n)} & \multicolumn{3}{c}{Average} \\
        \cmidrule(lr){2-6}\cmidrule(lr){7-10}\cmidrule(lr){11-13}
        Method & \multicolumn{1}{c}{\rotatebox{90}{Obs.}} & \multicolumn{1}{c}{\rotatebox{90}{Put back}} & \multicolumn{1}{c}{\rotatebox{90}{Rearr.}} & \multicolumn{1}{c}{\rotatebox{90}{Swap}} & \multicolumn{1}{c}{\rotatebox{90}{Swap T}} & \multicolumn{1}{c}{\rotatebox{90}{Batt.}} & \multicolumn{1}{c}{\rotatebox{90}{Rank.}} & \multicolumn{1}{c}{\rotatebox{90}{Cover}} & \multicolumn{1}{c}{\rotatebox{90}{Press}} & \multicolumn{1}{r}{M(1)} & \multicolumn{1}{r}{M(n)} & \multicolumn{1}{r}{All} \\
        \midrule
        \PI~\cite{physicalintelligence2025pi05} & 9 & 11 & 13 & 24 & 15 & 16 & 6 & 0 & 0 & 14.4 & 5.5 & 10.4 \\
        X-VLA~\cite{zheng2025xvla} & 9 & 18 & 13 & 16 & 3 & 26 & 1 & 2 & 0 & 11.8 & 7.2 & 9.8 \\
        \midrule
        Mem-0~\cite{chen2026rmbench} & 4 & 90 & 89 & 67 & 14 & 28 & 18 & 68 & 0 & 52.8 & 28.5 & 42.0 \\
        MemER~\cite{sridhar2025memer} & 7 & 0 & 17 & 14 & 7 & 27 & 0 & 6 & 0 & 9.0 & 8.2 & 8.7 \\
        MemoryVLA$_{\text{OpenVLA}}$~\cite{shi2025memoryvla} & 0 & 50 & 22 & 17 & 9 & 25 & 12 & 40 & 0 & 19.6 & 19.2 & 19.4 \\
        MemoryVLA$_{\text{QwenOFT}}$~\cite{shi2025memoryvla} & 2 & 81 & 53 & \cellcolor{tabr3}76 & 9 & 33 & 53 & 69 & 0 & 44.2 & 38.8 & 41.8 \\
        EventVLA~\cite{yang2026eventvla} & \cellcolor{tabr2}21 & \cellcolor{tabr3}95 & \cellcolor{tabr2}96 & \cellcolor{tabr2}96 & \cellcolor{tabr1}87 & 35 & \cellcolor{tabr3}81 & \cellcolor{tabr1}97 & 3 & \cellcolor{tabr2}79.0 & \cellcolor{tabr2}54.0 & \cellcolor{tabr2}67.9 \\
        \midrule
        ChainVLA~\cite{huang2026chainvla} & 11 & \cellcolor{tabr2}96 & 93 & 74 & 40 & -- & -- & -- & -- & 62.8 & -- & -- \\
        ReCAT~\cite{vanjani2026recat} & 4 & \cellcolor{tabr1}100 & \cellcolor{tabr2}96 & \cellcolor{tabr1}100 & \cellcolor{tabr3}64 & \cellcolor{tabr3}43 & \cellcolor{tabr1}99 & 50 & \cellcolor{tabr3}6 & \cellcolor{tabr3}72.8 & 49.5 & \cellcolor{tabr3}62.4 \\
        \midrule
        MemBodied~\cite{pala2026membodied} & -- & 38 & 94 & 50 & -- & 38 & 20 & -- & -- & -- & -- & -- \\
        DRAM~\cite{zhao2026dram} & \cellcolor{tabr3}15 & 65 & \cellcolor{tabr3}95 & \cellcolor{tabr1}100 & 45 & \cellcolor{tabr1}65 & 45 & \cellcolor{tabr3}70 & \cellcolor{tabr2}20 & 64.0 & \cellcolor{tabr3}50.0 & 57.8 \\
        \textbf{\ours (ours)} & \cellcolor{tabr1}30 & \cellcolor{tabr1}100 & \cellcolor{tabr1}100 & \cellcolor{tabr2}96 & \cellcolor{tabr2}82 & \cellcolor{tabr2}62 & \cellcolor{tabr2}82 & \cellcolor{tabr2}84 & \cellcolor{tabr1}70 & \cellcolor{tabr1}81.6 & \cellcolor{tabr1}74.5 & \cellcolor{tabr1}78.4 \\
        \bottomrule
    \end{tabularx}
\end{table}

\vspace{-3pt}\paragraph{MIKASA-Robo-VLA.}
As shown in \cref{tab:main_mikasa}, \ours reaches 85.6\,\% mean success, comparable to the regression-based $\mu$VLA result (84.0\,\%) and above all flow-matching baselines.
The clearest separation occurs on RC5 and RSC, which require matching an earlier visual cue to objects after rearrangement.
\ours reaches 96\,\% and 84\,\% on these tasks, respectively, compared with at most 24\,\% and 16\,\% for the other flow-matching memory policies.

\vspace{-3pt}\paragraph{RMBench.}
Across nine bimanual tasks, \ours achieves 78.4\,\% mean success, compared with 67.9\,\% for EventVLA, an explicit keyframe baseline (\cref{tab:main_rmbench}).
The gap is larger on tasks requiring multi-step state accumulation:
\ours reaches 74.5\,\% on $\mathrm{M}(n)$, compared with 54.0\,\% for EventVLA.
On single-cue recall tasks $\mathrm{M}(1)$, the corresponding rates are 81.6\,\% and 79.0\,\%.

\vspace{-3pt}\subsection{Real-World Manipulation}
\label{sec:exp_realworld}
We compare \ours with memory-free \PI, MemBodied~\cite{pala2026membodied}, and EventVLA~\cite{yang2026eventvla} on three bimanual AgileX tasks (\cref{fig:curves}).
\ours improves success across the tasks, while memory-free \PI and MemBodied remain near chance.
EventVLA performs well on layout restoration, where keyframes can preserve the return locations.
In color recall, however, the policy must match the remembered cue to a shuffled location within the narrow decision window, showing a failure pattern similar to simulation.
Pressing buttons requires tracking completed presses to decide when to stop, but the baselines sometimes press the wrong number of times or continue pressing without stopping.
Protocols and OOD robustness tests are provided in \appref{app:realworld}.

\begin{figure}[t]
    \centering
    \includegraphics[width=\linewidth]{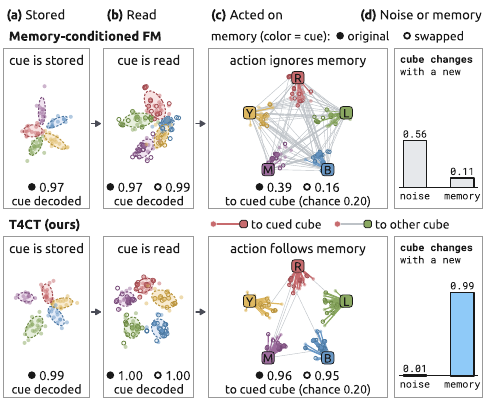}
    \caption{\textbf{Memory-to-action analysis on RC5.}
    We trace the cue from stored memory to action for \emph{memory-conditioned FM} and \ours.
    Each point represents a memory, colored by its cue and either original or swapped in from an episode with a different cue.
    Letters denote the five cube colors (\textbf{R}ed, \textbf{L}ime, \textbf{B}lue, \textbf{Y}ellow, \textbf{M}agenta).
    \textbf{(a,~b)}~Linear discriminant analysis (LDA) projections of the stored fast weights and the memory reads $m_t$, respectively.
    Numbers report linear-probe accuracy for the cue color.
    \textbf{(c)}~The cube that each memory's actions reach.
    Colored lines mark actions aligned with the memory's cue, and gray lines mark those that are not.
    Numbers report the fraction of aligned actions.
    \textbf{(d)}~Fraction of actions whose target cube changes when either the noise or the memory is replaced.}
    \label{fig:memrepr}
\end{figure}

\vspace{-5pt}
\subsection{Does Memory Guide Action Selection?}
\label{sec:exp_main}
To locate where memory-conditioned policies fail, we trace the cue on RC5 through three stages as illustrated in \cref{fig:memrepr}: stored fast weights, memory reads, and action selection.

\vspace{-5pt}\paragraph{Cue retention and action selection.}
For \emph{memory-conditioned FM}, linear probes recover the cued color from both the fast weights and the memory reads with 97\,\% accuracy (\cref{fig:memrepr}a,\,b).
Yet only 39\,\% of its actions reach the cued cube (chance: 20\,\%), compared with 96\,\% for \ours (\cref{fig:memrepr}c).
The baseline thus stores and retrieves the cue, but fails to translate it into action selection.

\vspace{-5pt}\paragraph{Counterfactual memory interventions.}
We next swap in a memory with a different cue while keeping the observation and initial noise fixed.
\ours follows the substituted cue in 95\,\% of actions, compared with 16\,\% for \emph{memory-conditioned FM} (open markers in \cref{fig:memrepr}c).
Resampling the noise instead changes the target cube in 56\,\% of actions for \emph{memory-conditioned FM}, but in only 1\,\% for \ours (\cref{fig:memrepr}d).
Conversely, resetting the fast weights before each call lowers \ours's success from 85.6\,\% to 33.2\,\% on MIKASA-Robo-VLA and from 78.4\,\% to 1.3\,\% on RMBench (\appref{app:source_interventions}), suggesting that manipulation relies on accumulated memory.

\providecolor{tabr1}{HTML}{C8E1FF}\providecolor{tabr2}{HTML}{E1F0FF}\providecolor{tabr3}{HTML}{F2F8FF}
\begin{table}[!t]
    \centering
    \caption{\textbf{Component ablations on MIKASA-Robo-VLA.} Success (\%) at 30K steps on the same evaluation seeds. Each variant removes or changes one part of \ours on the same backbone, data, and schedule. We evaluate 50 episodes.}
    \label{tab:ablation_main}
    \papertablefont
    \fontsize{7}{8.4}\selectfont
    \setlength{\tabcolsep}{1.6pt}
    \begin{tabularx}{\linewidth}{>{\raggedright\arraybackslash}Xrrrrrr}
        \toprule
        Method & RC5 & RSC & SGP & TIB & IM & Avg. \\
        \midrule
        \rowcolor{tabours}\ours (full) & \textbf{96} & \textbf{84} & 98 & \textbf{98} & \textbf{52} & \textbf{85.6} \\
        \midrule
        \textit{w/o} associative memory & 16 & 6 & 34 & 86 & 38 & 36.0 \\
        \textit{w/o} $\mathcal L_{\rm readout}$ & 20 & 22 & 90 & 90 & 42 & 52.8 \\
        \textit{w/o} MIF & \textbf{96} & 80 & 98 & 90 & 40 & 80.8 \\
        TBPTT over 1 call & 88 & 36 & 98 & 90 & 36 & 69.6 \\
        TBPTT over 4 calls & 92 & 74 & \textbf{100} & 96 & 46 & 81.6 \\
        TBPTT over 8 calls & 94 & 68 & 96 & 86 & 26 & 74.0 \\
        \midrule
        Memory-conditioned FM & 8 & 18 & 98 & 84 & 36 & 48.8 \\
        \bottomrule
    \end{tabularx}
\end{table}

\subsection{Ablation Studies}

\begin{figure}[t!]
    \centering
    \includegraphics[width=\linewidth]{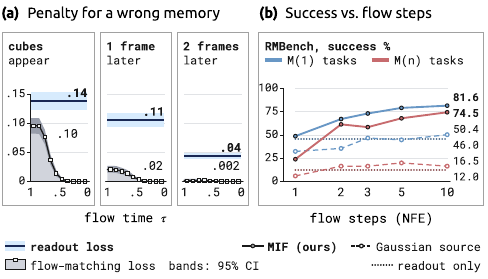}
    \caption{\textbf{Readout supervision and flow initialization.}
    \textbf{(a)}~Loss penalty $\Delta\mathcal L=\mathcal L_{\text{wrong}}-\mathcal L_{\text{original}}$ for a wrong-cue memory on early RC5 frames (3K training steps). Numbers report the readout penalty and peak flow-matching penalty over $\tau$. Bands show 95\,\% confidence intervals across episode resamples.
    \textbf{(b)}~RMBench success across flow steps (NFE) on $\mathrm{M}(1)$ and $\mathrm{M}(n)$ tasks, comparing MIF with an identical variant trained with a Gaussian source.}
    \label{fig:mif}
\end{figure}

\vspace{-5pt}\paragraph{Readout supervision.}
The \emph{w/o} $\mathcal L_{\mathrm{readout}}$ variant removes the readout loss and the stop-gradient on the flow source.
The flow-matching loss alone then trains both the readout and memory.
This variant reduces mean success on MIKASA-Robo-VLA from 85.6\,\% to 52.8\,\% (\cref{tab:ablation_main}).
As shown in \cref{fig:mif}a, we compare losses under original and wrong-cue memories on a Gaussian-source checkpoint at 3K training steps, at cube appearance and the next two frames.
The observation, target action, and noise are held fixed within each comparison.
The flow-matching penalty concentrates at high noise and rapidly approaches zero over the next two frames as movement reveals the selected target.
The readout penalty also decreases, but remains positive across these frames without receiving the interpolant.

\vspace{-5pt}\paragraph{Backpropagation window.}
Because retrieval reads $S_{t-1}$ before the current write, supervising the preceding memory write requires backpropagation across at least two calls.
Extending TBPTT from one to two calls raises mean success from 69.6\,\% to 85.6\,\% (\cref{tab:ablation_main}).
Four- and eight-call windows yield lower success despite propagating gradients through more memory writes.

\label{sec:exp_ablation}

\vspace{-3pt}\paragraph{Flow refinement and initialization.}
On MIKASA-Robo-VLA, readout supervision with a Gaussian flow source already reaches 80.8\,\% mean success.
On RMBench, where longer action chunks are executed, flow refinement raises success from 30.9\,\% for direct readout execution to 78.4\,\% on the same checkpoints (\cref{fig:mif}b).
Keeping all other components, including readout supervision, a variant trained from a Gaussian source reaches 35.3\,\% at ten integration steps.
Readout initialization also maintains higher success with fewer flow steps (NFE), with particularly large gains on $\mathrm{M}(n)$ tasks requiring accumulated state.
These results suggest that initializing from the memory-informed readout helps the flow preserve the intended action choice while refining its execution, yielding higher task success even at matched integration budgets.

\section{Discussion}
\label{sec:discussion}

\vspace{-3pt}
\paragraph{Conclusion.}
Efforts to equip VLA policies with episodic memory have largely focused on how to store and compress past observations. Our findings point to a complementary bottleneck: when memory is paired with a flow-matching decoder, a policy can retain episode context with high fidelity yet fail to act on it, because the noisy action interpolant and ongoing robot motion already reveal the target outside a brief decision window.
\ours investigates and addresses this mismatch by separating cue-dependent mode commitment from continuous trajectory refinement.
Supervising fast-weight associative memory through a direct action readout compels the memory projections to resolve the intended action without the interpolant shortcut, while initializing the flow from that readout refines the recalled choice into precise and continuous control.

\vspace{-3pt}
\paragraph{Limitations and future work.}
Our evaluation assumes static tabletop manipulation where task-relevant cues are unambiguously revealed early in the episode.
Real-world deployment, however, often violates both assumptions: in mobile manipulation, camera viewpoints shift continuously across rooms, and relevant cues may be partial, noisy, or absent without active exploration.
Under such conditions, maintaining a single point estimate can be brittle when past evidence remains uncertain.
Extending memory to mobile manipulation through uncertainty-aware memory updates~\cite{bui2026kdn} and active information gathering is an important direction for future work.

\section*{LLM Usage Statement}
We used Claude (Anthropic) and Codex (OpenAI) to assist with
manuscript refinement, experimental design, and implementation.
All content was reviewed and verified by the authors, who
take full responsibility for the submitted work.

\section*{Acknowledgement}
This work was supported by Institute of Information \& communications Technology Planning \& Evaluation (IITP, AI Computing Support Project for R\&D) grant funded by the Korea government(MSIT) (High-Performance Research AI Computing Infrastructure Support at the 2 PFLOPS Scale, RS-2026-25505492).

{
    \small
    \bibliographystyle{ieeenat_fullname}
    \bibliography{references}
}

\clearpage
\maketitlesupplementary
\appendix
\setcounter{figure}{0}\renewcommand{\thefigure}{S\arabic{figure}}\renewcommand{\theHfigure}{S\arabic{figure}}
\setcounter{table}{0}\renewcommand{\thetable}{S\arabic{table}}\renewcommand{\theHtable}{S\arabic{table}}
\setcounter{equation}{0}\renewcommand{\theequation}{S\arabic{equation}}\renewcommand{\theHequation}{S\arabic{equation}}

\noindent
This supplementary material first specifies the experimental setup, namely the benchmarks and evaluation protocol (\cref{app:benchmarks}), the baselines (\cref{app:baselines}), and the implementation of \ours (\cref{app:mif_recipe}).
We then derive the memory gain of \cref{sec:memory_unused} (\cref{app:memory_gain}) and the coupling-cost bound of \cref{sec:method_mif} (\cref{app:grounded_theory}).
The remaining sections provide the analyses behind \cref{sec:analysis,sec:exp_main,sec:exp_ablation}, covering a toy model of the interpolant shortcut (\cref{app:shortcut_toys}), its measurement on RC5 (\cref{app:shortcut_rc5}), what the current observation reveals (\cref{app:present}), and how trained policies use memory (\cref{app:policy_use}).

\section{Benchmarks and Evaluation Protocol}
\label{app:benchmarks}
\label{app:details}
\label{app:evaluation_protocol}

\begin{table*}[t]
    \centering\papertablefont
    \caption{\textbf{Tasks and what each one requires the policy to remember.} On MIKASA-Robo-VLA, the remembered item either keeps its \emph{place}, must be found again by appearance after the scene changes (\emph{match}), or is a \emph{motion}. RMBench kinds follow the benchmark, with $\mathrm{M}(1)$ tasks requiring one past observation and $\mathrm{M}(n)$ tasks depending on the outcomes of several trials.}
    \label{tab:benchmarks}
    \begin{tabularx}{\linewidth}{@{}ll>{\raggedright\arraybackslash}X@{}}
        \toprule
        Task & Kind & What the policy must carry forward \\
        \midrule
        \multicolumn{3}{@{}l}{\emph{MIKASA-Robo-VLA, one policy for five tasks}} \\
        ShellGamePush (SGP) & place & which of three cups covered the ball before the cups closed, to push that cup to the goal \\
        TakeItBack (TIB) & place & the cube's initial region, to push it back there after it has been pushed to a goal \\
        RememberColor5 (RC5) & match & the color of a cue cube that disappears, to touch the cube of that color among five reshuffled cubes \\
        RememberShapeAndColor3x3 (RSC) & match & the shape and color of a cue object, to touch the matching one of nine objects that combine three shapes and three colors \\
        InterceptMedium (IM) & motion & the motion of a launched ball, to intercept it and push it into the goal \\
        \midrule
        \multicolumn{3}{@{}l}{\emph{RMBench, one policy per task}} \\
        Put back & $\mathrm{M}(1)$ & which of four mats held the block, to return it there after moving it to the center and pressing a button \\
        Rearrange & $\mathrm{M}(1)$ & which block started on a mat, to move it between the mats after the other block has filled the empty mat \\
        Swap & $\mathrm{M}(1)$ & the trays of two blocks, to swap them through a third tray one block at a time \\
        Swap T & $\mathrm{M}(1)$ & the positions and orientations of two T-shaped blocks, to swap their poses \\
        Observe & $\mathrm{M}(1)$ & the object on a shelf before a screen hides it, to pick its twin among five objects on the table \\
        Battery & $\mathrm{M}(n)$ & which battery orientations have already been tried, to find the one that turns the needle \\
        Ranking & $\mathrm{M}(n)$ & which arrangements of three cubes have failed, to keep rearranging and pressing the button until one succeeds \\
        Cover & $\mathrm{M}(n)$ & which lid covers which colored block, to uncover them in the order red, green, blue \\
        Press & $\mathrm{M}(n)$ & how many times each button has been pressed, to press the left and middle buttons as many times as the two numbers on the table and then confirm \\
        \bottomrule
    \end{tabularx}
\end{table*}

\paragraph{MIKASA-Robo-VLA.}
MIKASA-Robo-VLA~\cite{cherepanov2026mikasarobovla,cherepanov2025mikasa} poses tabletop tasks for a Franka arm in which the information needed for success is visible only early in the episode (\cref{tab:benchmarks}).
We train a policy on the five tasks selected by $\mu$VLA~\cite{cherepanov2026muvla} from 250 demonstrations per task, with a fixed top camera and a wrist camera, the 7-D robot state, and 7-D actions (\ie, a 6-D end-effector pose delta and a gripper command).
Episodes end after 25 steps on RC5 and RSC, 30 steps on SGP, and 60 steps on TIB and IM.
Each policy is evaluated on the same 50 environment seeds per task.

\paragraph{RMBench.}
RMBench~\cite{chen2026rmbench} poses bimanual tasks in simulation whose next movement depends on earlier observations or on the outcomes of earlier trials.
Following prior work~\cite{chen2026rmbench,pala2026membodied}, we train one policy per task from its 50 demonstrations, with a head camera and two wrist cameras, resized with padding to $224\times224$, and the 14-D joint state of both arms. We evaluate our policy on 50 episodes for each task.
All tasks use the same observation-driven memory, action-readout supervision, and flow-refinement procedure, with the training settings listed in \cref{tab:hparams}.

\subsection{Real-World Tasks}
\label{app:realworld}

\begin{figure*}[t]
    \centering
    \includegraphics[width=\linewidth]{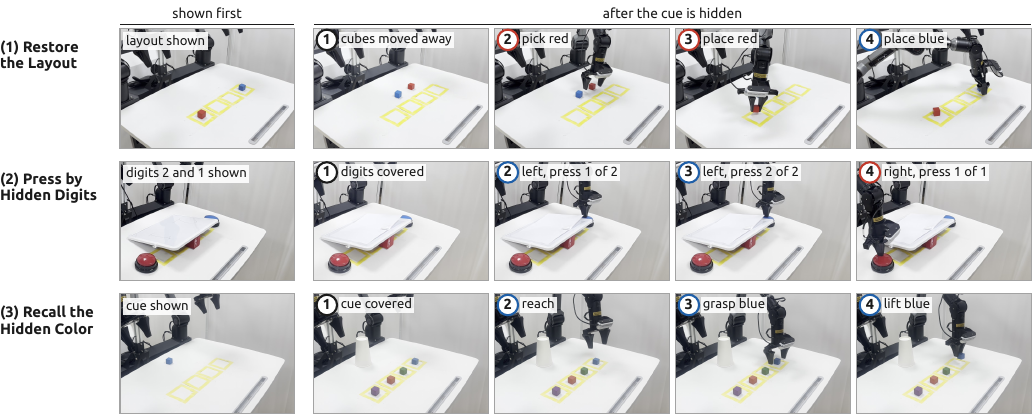}
    \caption{\textbf{Episodes of the three real-world tasks.} Each row shows one episode of a task. The leftmost frame shows the scene while the cue is visible, frame 1 shows the scene once the cue has been moved away or covered, and frames 2 to 4 show the robot acting. Frames come from a camera on a fixed stand beside the robot.}
    \label{fig:task_episodes}
\end{figure*}

\paragraph{Setup.}
The real-world experiments (\cref{fig:curves}) use an AgileX Cobot Magic robot with two Piper arms, a front camera, and two wrist cameras.
All demonstrations are collected by teleoperation.
The four policies in \cref{tab:realworld} are trained under the same conditions.
Each starts from the same pretrained \PI checkpoint, is trained on the same demonstrations of each task, takes the three camera images, the 14-D joint state, and the instruction as input, and predicts chunks of 50 actions.
\ours and \PI are trained with the RMBench settings of \cref{tab:hparams}, and MemBodied and EventVLA add their memory to \PI as described in their papers.

\paragraph{Tasks.}
\Cref{fig:task_episodes} shows one episode of each of the three tasks, one per row, seen from a camera beside the robot.
In \emph{Restore the Layout} (row 1), a red and a blue cube are placed on two of four identical taped slots (12 layouts, chance $1/12$).
A hand then moves both cubes to the middle of the table, and the left arm must return the red cube and the right arm the blue cube, given only the instruction ``put the two cubes back where they were''.
We record 96 demonstrations, eight per layout.
In \emph{Press by Hidden Digits} (row 2), two digits from 1 to 3 are shown on a whiteboard for about two seconds before a hand covers them (9 combinations, chance $1/9$).
The left arm then presses the left button as many times as the left digit, and the right arm presses the right button as many times as the right digit, so the policy must retain both the digits and the number of presses so far.
We record 45 demonstrations, five per combination.
In \emph{Recall the Hidden Color} (row 3), a cue cube is shown in the middle of the table and covered with a cup.
Four cubes of different colors are then placed on the four slots in a new arrangement (24 arrangements, chance $1/4$), and the arm on the side of the matching cube picks it.
We record 48 demonstrations.

\paragraph{Evaluation.}
\Cref{tab:realworld} reports the success rates plotted in \cref{fig:curves}.
We run 24 trials for \emph{Restore the Layout} (each layout twice), 18 trials for \emph{Press by Hidden Digits} (each digit pair twice), and 20 trials for \emph{Recall the Hidden Color}.
A Restore trial succeeds when both cubes return to their original slots, a Press trial when each arm presses its button the shown number of times, and a Recall trial when the arm reaches the cube of the cued color.

\begin{table}[t]
    \centering\papertablefont
    \setlength{\tabcolsep}{6pt}
    \caption{\textbf{Real-world success.} Success rate (\%) of each policy on the three tasks of \cref{fig:task_episodes}, over 24, 18, and 20 trials, respectively.}
    \label{tab:realworld}
    \begin{tabularx}{\linewidth}{>{\raggedright\arraybackslash}Xccc}
        \toprule
        Method & Restore & Press & Recall \\
        \midrule
        \PI~\cite{physicalintelligence2025pi05} & 4.2 & 0.0 & 20.0 \\
        MemBodied~\cite{pala2026membodied} & 16.7 & 5.6 & 25.0 \\
        EventVLA~\cite{yang2026eventvla} & 75.0 & 22.2 & 25.0 \\
        \midrule
        \rowcolor{tabours}\ours & \textbf{95.8} & \textbf{72.2} & \textbf{95.0} \\
        \bottomrule
    \end{tabularx}
\end{table}

\begin{figure*}[t]
    \centering
    \includegraphics[width=\linewidth]{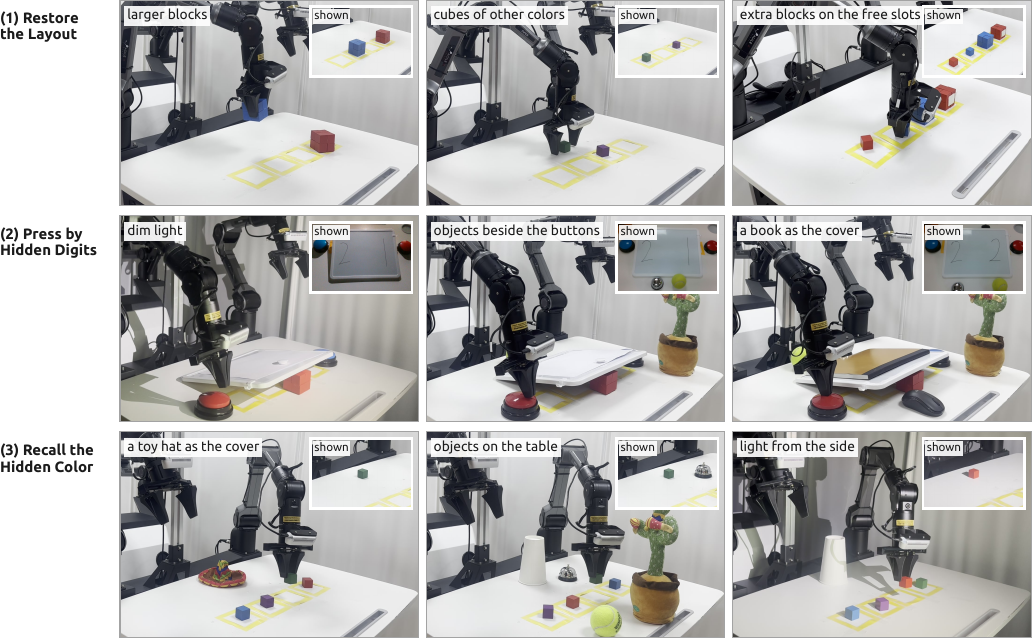}
    \caption{\textbf{Out-of-distribution trials on the real robot.} Each frame shows a successful trial of \ours under a condition absent from the demonstrations. The inset of each frame shows what was shown before it was hidden. Insets of the Press frames come from the robot's front camera, and all other frames from the camera beside the robot.}
    \label{fig:ood_examples}
\end{figure*}

\paragraph{Out-of-distribution trials.}
We also run \ours under conditions that do not appear in the demonstrations, using the policies of \cref{tab:realworld} without further training.
Each condition is run for a small number of trials, so we present them as qualitative examples rather than as success rates.
\Cref{fig:ood_examples} shows one successful trial for each of nine conditions.
In \emph{Restore the Layout} (row 1), the cubes are replaced by larger blocks or by cubes of other colors, or extra blocks occupy the free slots.
In \emph{Press by Hidden Digits} (row 2), the light is dimmed, unrelated objects are placed beside the buttons, or the digits are covered with a book.
In \emph{Recall the Hidden Color} (row 3), a toy hat covers the cue in place of the cup, unrelated objects are placed on the table, or the table is lit from the side.
In each of these trials, the robot acts on what it was shown before the scene changed.

\begin{figure}[t]
    \centering
    \includegraphics[width=\linewidth]{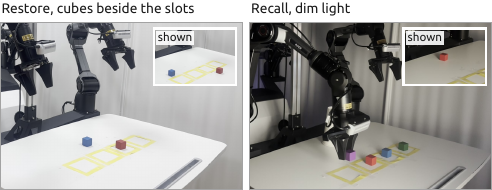}
    \caption{\textbf{Failed out-of-distribution trials.} Two trials of \ours that fail, shown as in \cref{fig:ood_examples}.}
    \label{fig:ood_failures}
\end{figure}

\paragraph{Failure cases.}
Not every such trial succeeds, and \cref{fig:ood_failures} shows two failed trials.
In \emph{Restore the Layout}, when the cubes start beside the taped slots instead of on them, the left arm places the red cube on a slot near its original position, and the right arm leaves the blue cube untouched.
In \emph{Recall the Hidden Color} under dim light, the arm grasps the purple cube next to the cued red one.
Cube positions off the slots and dim light are not covered by the demonstrations, and we expect that handling them would require demonstrations with more varied placements and lighting.

\section{Baselines}
\label{app:baselines}

\paragraph{Memory-conditioned FM.}
This baseline follows the design of \cref{sec:memcond}, in which the reads of the test-time memory are given to the action expert as additional condition tokens alongside the observation.
It shares the backbone and the fast-weight memory of \ours (\cref{app:mif_recipe}) and is trained from the same \PI checkpoint on the same data.
Its memory is trained end to end through the flow-matching loss.
It uses neither the readout loss nor the readout source of \ours, and generates actions from Gaussian noise as in standard conditional flow matching.

\paragraph{Compared methods.}
We briefly describe the methods compared in \cref{tab:main_mikasa,tab:main_rmbench}, in the order of the memory families of the tables.
Among memory-free policies, \PI~\cite{physicalintelligence2025pi05} pairs a PaliGemma vision-language model with a flow-matching action expert and acts on the current observation.
OpenVLA-OFT~\cite{kim2025openvlaoft} fine-tunes OpenVLA to regress action chunks in parallel.
X-VLA~\cite{zheng2025xvla} is a flow-matching transformer policy that adapts to each embodiment through soft prompts.

Among policies that store past frames or tokens, Decision Transformer (DT)~\cite{chen2021decision} predicts actions from a context window of past observations and actions.
MemoryVLA~\cite{shi2025memoryvla} maintains a bank of perceptual and cognitive tokens and retrieves from it to condition action generation.
MemER~\cite{sridhar2025memer} uses a high-level vision-language model to select the keyframes to keep and to instruct a low-level policy.
Mem-0~\cite{chen2026rmbench} is the modular policy with explicit memory components introduced together with RMBench.
EventVLA~\cite{yang2026eventvla} records event-driven visual keyframes across long horizons.

Among recurrent policies, RATE~\cite{cherepanov2026rate} passes recurrent memory embeddings between consecutive context windows of a transformer.
ELMUR~\cite{cherepanov2025elmur} attaches an external memory to each transformer layer and updates it as the episode proceeds.
$\mu$VLA~\cite{cherepanov2026muvla} carries recurrent memory tokens through a VLA from one step to the next.
ChainVLA~\cite{huang2026chainvla} chains successive policy queries through a unified execution state.
ReCAT~\cite{vanjani2026recat} uses a structured recurrent memory designed for remembering, counting, and timing.

Among test-time associative-memory policies, MemBodied~\cite{pala2026membodied} and DRAM~\cite{zhao2026dram} accumulate the episode history in a recurrent associative memory and condition action generation on what is retrieved from it.
In \cref{tab:main_mikasa}, \PI + MemoryVLA and \PI + $\mu$VLA denote the respective memory applied to \PI.

In \cref{tab:main_mikasa}, \PI and the three memory baselines built on it are trained by us from the same \PI checkpoint, on the same demonstrations, and for the same number of training steps as \ours.
The remaining rows of \cref{tab:main_mikasa,tab:main_rmbench} report published results, and dashes in \cref{tab:main_rmbench} mark tasks that the original paper does not report.

\section{Implementation Details}
\label{app:mif_recipe}

\paragraph{Backbone.}
\ours starts from the pretrained \PI base checkpoint~\cite{physicalintelligence2025pi05}.
The VLM is adapted with rank-16 LoRA, while the memory modules, the readout decoder, and the action expert are fully trained.

\paragraph{Memory.}
Each of the $L$ Gemma layers carries one $256\times256$ fast weight, written once per call from the keys and values of the visual tokens by \cref{eq:mif_inner}.
With $\Lambda_t=\operatorname{diag}(w_{t,i})$ and $\|X\|_{\Lambda}^2=\operatorname{tr}(X^\top\Lambda X)$, the update has the closed form
\begin{equation}
 S_t=\gamma_t S_{t-1}+(I+K_t^\top\Lambda_t K_t)^{-1}K_t^\top\Lambda_t\,(V_t-K_t\gamma_t S_{t-1}),
 \label{eq:mif_inner_closed}
\end{equation}
which requires one linear solve per layer and call.
Keys and values are normalized, and the token weights $\Lambda_t$ and the retention $\gamma_t$ are learned.
Eight learned queries per layer read the previous state $S_{t-1}$ after attending to the keys of the current call.
On RMBench, whose tasks depend on the order and number of earlier events, each stored key and value also carries a sinusoidal code of the call index, so that the memory records when an observation was written.
A read thus returns the content together with the time at which it was written, and a second read retrieves what was written a learned number of calls after that time and adds it to the first.

\paragraph{Readout decoder.}
The readout decoder $D_\psi$ has three layers of width 256 with four heads.
Its learned action queries attend to the visual tokens of the current frame, the state and instruction tokens, and the memory tokens $m_t$.
A detached copy of $m_t$ also conditions the action expert (\cref{eq:mif_flow_loss}).

\paragraph{Training.}
A training sample is one policy call, made of the current observation $o_t$, the memory $S_{t-1}$ read before this call's write, and the demonstrated chunk $a_t^\star$ of $H$ actions.
The source is $x_1=\stopgrad(\hat a_t)+\sigma\epsilon$ with $\sigma=0.1$, and the flow time follows the shifted beta distribution of \PI.
The flow and readout losses have unit weight.
On RMBench, an auxiliary classifier on the memory additionally predicts the movement-level instruction annotated in the demonstrations.
The readout loss trains the memory and the readout decoder and reaches earlier writes within each truncated window of consecutive calls, whereas the flow loss trains the expert and the VLM adapters, as the source and the memory tokens are detached.

\paragraph{Execution.}
\label{app:grounded_contract}
At each call, the policy reads $S_{t-1}$, forms the readout, integrates the flow from the readout-centered source with ten Euler steps, and then writes the current frame.
All fast weights are reset at episode boundaries.
At deployment, only the fast weights change, through the observation-only regression of \cref{eq:mif_inner}.

\paragraph{Hyperparameters.}
\Cref{tab:hparams} lists the main training settings.
All runs use AdamW, and images are resized to $224\times224$ and augmented with random crops and color jitter.

\begin{table}[t]
    \centering\papertablefont
    \setlength{\tabcolsep}{4pt}
    \caption{\textbf{Training settings.} Learning rates, in units of $10^{-4}$, are listed for the action expert, the memory, the readout decoder, and the LoRA adapters. Schedules decay by a cosine to 10\,\% of the peak rate.}
    \label{tab:hparams}
    \begin{tabular}{@{}lll@{}}
        \toprule
        Setting & MIKASA-Robo-VLA & RMBench \\
        \midrule
        Cameras & 2 & 3 \\
        State and action size & 7 & 14 \\
        Chunk $H$ & 8 & 50 \\
        Actions run per call & 1 & 50 \\
        Learning rates & 1, 1, 3, 1 & 0.25, 1, 3, 1 \\
        Weight decay, clip & 0.01, 1.0 & 0.01, 1.0 \\
        LoRA rank, $\alpha$ & 16, 16 & 16, 16 \\
        Read queries per layer & 8 & 8 \\
        Euler steps & 10 & 10 \\
        \bottomrule
    \end{tabular}
\end{table}

\paragraph{Ablation variants.}
The variants in \cref{tab:ablation_main} share the backbone, data, training schedule, and evaluation episodes.
\emph{W/o associative memory} resets the fast weights before every call in both training and evaluation, whereas the memory reset of \cref{app:source_interventions} clears the memory of the full checkpoint only at evaluation.
\emph{W/o $\mathcal L_{\rm readout}$} removes the readout loss and the stop-gradient on the source in \cref{eq:mif_flow_loss}, so that the flow loss trains the decoder and the memory.
\emph{W/o MIF} keeps readout supervision, but trains and executes the flow from $\mathcal N(0,I)$.

\section{Memory and the Flow-Matching Loss}
\label{app:memory_gain}

This section quantifies how much a memory can lower the flow-matching loss.
We first show that, at every flow time $\tau$, the reduction equals how far the memory moves the best estimate of the demonstrated action, which gives the memory gain $\mathcal G(\tau)$ of \cref{eq:decision_gain}.
We then show that, when the memory selects between two separated action chunks, this gain is confined to a narrow range of flow times near pure noise, which gives the bound of \cref{eq:gain_decay}.

\paragraph{Velocity prediction is action prediction.}
Let $O$, $M$, and $A\in\R^{n}$ with $n=Hd_a$ denote the observation, the retrieved memory, and the demonstrated action chunk, with $\E\|A\|^2<\infty$, and let $\epsilon\sim\calN(0,I)$ be independent of them.
At flow time $\tau\in(0,1]$, the network receives the interpolant $X_\tau=(1-\tau)A+\tau\epsilon$ and regresses the velocity $U=\epsilon-A$.
Because the interpolant is given, the noise can be eliminated, which leaves
\begin{equation}
 U=\frac{X_\tau-A}{\tau},
 \label{eq:velocity_is_action}
\end{equation}
where the action is the only unknown.
A velocity prediction $h$ therefore corresponds to an action estimate $\hat a=X_\tau-\tau h$, and its squared error is $\|U-h\|^2=\|A-\hat a\|^2/\tau^2$.
In other words, flow matching at time $\tau$ asks the network to recover the demonstrated action from a noisy copy of it, and weighs the error by $1/\tau^2$.

\paragraph{The memory gain.}
The best estimate of the action in mean squared error is the posterior mean, so the smallest velocity risks without and with memory are
\begin{equation}
 R_0(\tau)=\frac{\E\|A-\mu_0\|^2}{\tau^2},\qquad
 R_m(\tau)=\frac{\E\|A-\mu_m\|^2}{\tau^2},
 \label{eq:conditional_risks}
\end{equation}
with $\mu_0=\E[A\mid X_\tau,O]$ and $\mu_m=\E[A\mid X_\tau,O,M]$.
To compare them, we split the error of the memory-free estimate as $A-\mu_0=(A-\mu_m)+(\mu_m-\mu_0)$.
The first part is what remains unknown even with the memory and has zero mean given $(X_\tau,O,M)$, whereas the second part is a function of $(X_\tau,O,M)$.
The two parts are therefore uncorrelated, their squared norms add, and
\begin{equation}
 \mathcal G(\tau)=R_0(\tau)-R_m(\tau)
 =\frac{1}{\tau^2}\,\E\|\mu_m-\mu_0\|^2\;\geq0,
 \label{eq:general_memory_gain_proof}
\end{equation}
which is \cref{eq:decision_gain}.
The identity makes no assumption on the action distribution or on the quality of the memory.

\paragraph{Reading the identity.}
The memory lowers the loss only to the extent that it moves the estimate of the action, that is, only where $\mu_m$ differs from $\mu_0$.
Two other inputs can provide the same information.
The current observation is one of them, and if what the memory indicates can already be read from $O$, then $\mu_m=\mu_0$ and the gain is zero at every $\tau$, which is the case examined in \cref{app:present}.
The interpolant is the other, and its effect depends on $\tau$.
At $\tau=1$ the interpolant is pure noise and carries no information on the action, so that
\begin{equation}
 \mathcal G(1)=\E\big\|\E[A\mid O,M]-\E[A\mid O]\big\|^2
 \label{eq:memory_gain_source}
\end{equation}
is the full value of the memory for predicting the action from the scene.
As $\tau$ decreases, $X_\tau$ becomes an increasingly accurate copy of $A$, and the memory-free estimate $\mu_0$ can read from it what the memory would have provided.
\Cref{fig:shortcut} shows this gain measured on RC5, with the estimation described in \cref{app:shortcut_rc5,app:present}, and the next subsection bounds how quickly it vanishes.

\subsection{A Two-Mode Decision Window}
\label{app:mif_theory}

Consider the simplest decision that requires memory.
Under the same observation, the demonstrated chunk is one of two distinct chunks $a^+$ and $a^-$ with equal probability, and the memory tells which.
Given the choice, the interpolant is Gaussian, $X_\tau\mid A=a^\pm\sim\calN\big((1-\tau)a^\pm,\tau^2I\big)$.
The two choices thus appear as two clouds of width $\tau$ whose centers lie $(1-\tau)\|a^+-a^-\|$ apart (top of \cref{fig:shortcut}).
Near $\tau=1$ the clouds coincide, and only the memory can tell the choices apart.
As $\tau$ decreases, the centers move apart while the clouds shrink, until the position of $X_\tau$ alone reveals the choice.
How well the clouds are separated is measured by the ratio of half the distance between the centers to the noise scale,
\begin{equation}
 \rho_\tau=\frac{(1-\tau)\,\|a^+-a^-\|}{2\tau},
 \label{eq:mode_snr}
\end{equation}
and the following result bounds the memory gain in terms of this ratio.

\medskip\noindent\textbf{Proposition 1.}
\emph{Let $a^+\neq a^-$ be two equally likely action chunks under the same observation, and let the memory reveal which one is demonstrated. Then, for every $\tau\in(0,1]$,}
\begin{equation}
 \frac{\mathcal G(\tau)}{\mathcal G(1)}
 \;\leq\;\frac{4}{\tau^2}\,\Phi(-\rho_\tau)
 \;\leq\;\frac{2}{\tau^2}\,e^{-\rho_\tau^2/2},
 \label{eq:vector_gain_bound}
\end{equation}
\emph{where $\Phi$ is the standard normal distribution function. The second bound is \cref{eq:gain_decay}.}
\medskip

\paragraph{Proof.}
Because the memory reveals the choice, $\mu_m=A$, and \cref{eq:general_memory_gain_proof} reduces to
\begin{equation}
 \mathcal G(\tau)=\frac{1}{\tau^2}\,\E\|A-\mu_0\|^2 .
 \label{eq:mif_gain_proof}
\end{equation}
The gain is thus the error of the best guess of the action that can be made without the memory.
No particular guess can do better than $\mu_0$, so we bound this error by that of a simple one.
Let $\hat A$ be the chunk whose cloud center $(1-\tau)a^\pm$ is nearer to $X_\tau$.
This guess is either correct, with no error, or selects the other chunk, with error $\|a^+-a^-\|^2$, so that
\begin{equation}
\begin{aligned}
 \E\|A-\mu_0\|^2&\leq\E\|A-\hat A\|^2\\
 &=\|a^+-a^-\|^2\,\Pr(\hat A\neq A).
\end{aligned}
 \label{eq:nearest_mode_error}
\end{equation}
The guess is wrong when the noise carries $X_\tau$ across the plane halfway between the two centers.
Along the direction $a^+-a^-$, each center lies at distance $(1-\tau)\|a^+-a^-\|/2$ from this plane and the noise has standard deviation $\tau$, which gives $\Pr(\hat A\neq A)=\Phi(-\rho_\tau)$.
At $\tau=1$ the interpolant is independent of the choice, the best guess is the midpoint of the two chunks, and $\mathcal G(1)=\|a^+-a^-\|^2/4$.
Dividing gives the first bound, and the Gaussian tail inequality $\Phi(-r)\leq\tfrac12e^{-r^2/2}$ for $r\geq0$ gives the second. \hfill$\square$

\paragraph{How narrow the window is.}
As $\tau$ decreases, the factor $1/\tau^2$ grows only polynomially, whereas $e^{-\rho_\tau^2/2}$ falls faster than any power of $\tau$, so the exponential term decides where the gain is located.
The clouds begin to separate when $\rho_\tau=1$, which with the half separation $d=\|a^+-a^-\|/2$ occurs at
\begin{equation}
 \tau_*=\frac{d}{1+d}.
 \label{eq:crossover}
\end{equation}
Above $\tau_*$ the choice is still hidden in the noise, and below it the bound falls rapidly.
For $d=3$, as in the toy model of \cref{app:shortcut_toys}, $\tau_*=0.75$ and the bound is already below $0.09$ at $\tau=0.5$.
The window $(\tau_*,1]$ has width $1/(1+d)$ and therefore narrows as the two chunks move apart.
Long action chunks make this separation large.
If the two choices differ by $\delta$ in each of $k$ coordinates, then $d=\delta\sqrt{k}/2$, so a decision that changes an entire reaching motion places $\tau_*$ close to one, while coordinates on which the choices agree do not enter $d$.
During training, frames and flow times are sampled, so the loss sees the gain averaged over both, and a gain confined to a few frames and to a narrow range of $\tau$ contributes little to this average.

\paragraph{Reference noise level.}
In the two-mode case, the action covariance has the single nonzero eigenvalue $d^2$.
For general action distributions we therefore replace $d^2$ by the largest eigenvalue $\Lambda$ of the action covariance, in line with the speciation criterion of Biroli \etal~\cite{biroli2024dynamical}, and use
\begin{equation}
 \tau_s=\frac{\sqrt\Lambda}{1+\sqrt\Lambda}
 \label{eq:tau_s}
\end{equation}
as a reference level, which coincides with $\tau_*$ for two modes.
\Cref{app:shortcut_toys,app:shortcut_rc5} use $\tau_s$ to locate the window in a toy model and on RC5.

\section{Starting the Flow at a Readout}
\label{app:grounded_theory}
\label{app:source_density}

This section examines what changes when the flow starts from the readout instead of Gaussian noise.
We first show that the readout error is, up to a constant, the mean squared distance that the flow has to cover, and that this distance bounds both the transport cost and the part of the flow loss that no expert can remove.
We then show that flow matching remains valid for such a source, although the source now depends on the scene and on the memory.

\subsection{The Readout Error Bounds the Source Cost}
\label{app:mif_source}

\paragraph{The distance the flow has to cover.}
Fix a context $c=(o,S)$ made of the current observation and the memory state, and let $A$ follow the demonstrated actions for this context.
The source sample is $X_1=\hat a+\sigma\epsilon$, where $\hat a=D_\psi(o,m)$ is the readout and $\epsilon\sim\calN(0,I)$ is independent of $A$.
Training pairs each source sample with a demonstrated action, and the flow learns to carry the former to the latter along a straight line.
The mean squared length of these lines is the coupling cost.
Since $X_1-A=(\hat a-A)+\sigma\epsilon$ and the noise is independent with zero mean, the cross term vanishes and
\begin{equation}
 \E\big[\|X_1-A\|^2\mid c\big]
 =\E\big[\|\hat a-A\|^2\mid c\big]+\sigma^2n,
 \label{eq:mif_cost_proof}
\end{equation}
whose average over contexts is \cref{eq:mif_coupling}, with the readout loss as the first term and a constant as the second.
For comparison, a standard Gaussian source that is independent of the action has cost $n+\E\|A\|^2$, at least one per action coordinate, whereas the readout source adds only $\sigma^2=0.01$ per coordinate to the readout error.

\paragraph{What the cost bounds.}
The first quantity is the squared 2-Wasserstein distance between the source law $q_\psi(\cdot\mid c)$ and the demonstrated law $p_{\rm data}(\cdot\mid c)$.
This distance is the smallest mean squared displacement over all ways of pairing samples of the two laws, and the pairing used in training is one of them, so that
\begin{equation}
\begin{aligned}
 &W_2^2\big(q_\psi(\cdot\mid c),\,p_{\rm data}(\cdot\mid c)\big)\\
 &\qquad\leq\E\big[\|X_1-A\|^2\mid c\big].
\end{aligned}
 \label{eq:w2_bound}
\end{equation}
The second quantity is the irreducible flow loss.
With the interpolant $X_\tau=(1-\tau)A+\tau X_1$, the expert regresses the velocity $X_1-A$ from $X_\tau$ and its conditioning $b$, which is the context or a part of it.
The smallest attainable risk at flow time $\tau$ is the variance that remains in this velocity once $X_\tau$ and $b$ are known.
A variance never exceeds the corresponding mean square, which the trivial prediction of zero velocity attains, so that for every $\tau$
\begin{equation}
\begin{aligned}
 &\E\big\|(X_1-A)-\E[X_1-A\mid X_\tau,b]\big\|^2\\
 &\qquad\leq\E\|X_1-A\|^2 .
\end{aligned}
 \label{eq:irreducible_bound}
\end{equation}
Because $X_1-A=(X_\tau-A)/\tau$, the left-hand side equals $\E\|A-\E[A\mid X_\tau,b]\|^2/\tau^2$, the error of the best action estimate weighed as in \cref{app:memory_gain}.
Integrating over flow time gives the branching cost of WarmPrior~\cite{kang2026warmprior}, written here with $\tau=1$ at the source,
\begin{equation}
\begin{aligned}
 \mathcal B&=\int_0^1\frac{\E\|A-\E[A\mid X_\tau,b]\|^2}{\tau^2}\,d\tau\\
 &\leq\E\|X_1-A\|^2,
\end{aligned}
 \label{eq:branching_cost}
\end{equation}
and the same bound holds under any other flow-time distribution.
Together with \cref{eq:mif_cost_proof}, lowering the readout loss tightens both bounds.
Intuitively, when every source sample already lies next to its target, the velocity left to regress is small and nearly unambiguous, which is consistent with the behavior at few integration steps in \cref{fig:mif}b.

\paragraph{Where the memory enters.}
The readout loss is minimized by the conditional mean of the action, and the memory lowers this minimum by
\begin{equation}
\begin{aligned}
 &\E\|A-\E[A\mid o]\|^2-\E\|A-\E[A\mid o,S]\|^2\\
 &\qquad=\E\big\|\E[A\mid o,S]-\E[A\mid o]\big\|^2,
\end{aligned}
 \label{eq:mif_context_value}
\end{equation}
by the same decomposition as in \cref{eq:general_memory_gain_proof}.
The right-hand side is the memory gain at pure noise, $\mathcal G(1)$ of \cref{eq:memory_gain_source}.
The readout objective therefore exposes the full value of the memory at every training step, independently of the sampled flow time, whereas the flow loss exposes it only inside the window of \cref{app:mif_theory}.
The source then passes the choice made by the readout on to the flow.

\subsection{Conditional Flow Matching from a Readout}
\label{app:conditional_transport}

The source of \ours is neither a fixed Gaussian nor independent of the target, since its center $\hat a$ is predicted from the same scene and memory that determine the demonstrated action.
Flow matching requires neither property.
It needs pairs of source and target samples joined by straight lines, and regressing the velocities of these lines yields a field that moves the whole source distribution onto the target distribution~\cite{lipman2023flowmatching,albergo2024couplings}.
Concretely, fix the expert conditioning $b=(o,m)$, let $(A,\hat a)$ follow the joint law of demonstrations and readouts given $b$, and set $X_1=\hat a+\sigma\epsilon$ and $X_\tau=(1-\tau)A+\tau X_1$.
The least-squares optimum of the velocity regression is
\begin{equation}
 v^\star(x,\tau;b)=\E[X_1-A\mid X_\tau=x,\,b].
 \label{eq:conditional_velocity}
\end{equation}
Since $X_\tau$ moves with velocity $X_1-A$, any smooth, compactly supported test function $f$ satisfies
\begin{equation}
\begin{aligned}
 \frac{d}{d\tau}\E[f(X_\tau)\mid b]
 &=\E[\nabla f(X_\tau)^\top(X_1-A)\mid b]\\
 &=\E[\nabla f(X_\tau)^\top v^\star(X_\tau,\tau;b)\mid b],
\end{aligned}
 \label{eq:continuity}
\end{equation}
where the second line replaces the velocity of each pair by its average at the point $X_\tau$.
The distribution of $X_\tau$ therefore evolves as if every sample followed $v^\star$, which is the continuity equation in weak form.
When the ODE is well posed, integrating $v^\star$ from $\tau=1$ to $0$ carries the source law $q(X_1\mid b)$ to the demonstrated law $p^\star(A\mid b)$.

This statement concerns the distribution as a whole.
Whether each sample keeps the target chosen by its readout depends on the pairing, which the readout loss shapes by keeping source and target close (\cref{app:mif_source}).
During training the source changes as the readout improves, and the stop-gradient in \cref{eq:mif_flow_loss} lets the flow fit the current source without altering it.

\paragraph{The readout selects the target and the flow refines the motion.}
Measurements on the trained policy are consistent with this division of roles.
On RSC, the readout and the flow started from it aim at the same object in 94\,\% of decision calls.
When started from $\calN(0,I)$ instead, the same expert succeeds in none of 50 RSC episodes, which suggests that it relies on the readout for the target.
The flow in turn contracts the source spread of 0.1 to 0.039 on RSC and to 0.0067 on RMBench Swap T.

\section{Interpolant Shortcut in a Toy Model}
\label{app:shortcut_toys}

\begin{figure}[t]
    \centering
    \includegraphics[width=\linewidth]{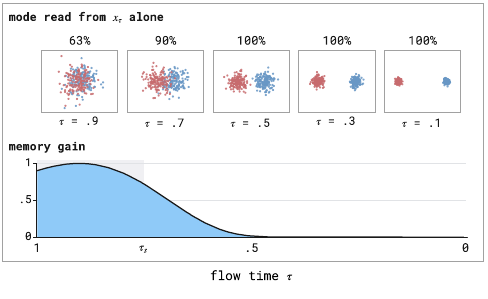}
    \caption{\textbf{Interpolant shortcut in a two-mode toy model.} The top row shows the interpolants $x_\tau$ of the two modes at five flow times, with the accuracy of identifying the mode from $x_\tau$ alone. The curve below shows the memory gain over flow time, scaled to its maximum, of which 76\,\% lies above $\tau_s=0.75$ (shaded).}
    \label{fig:toyA}
\end{figure}

\begin{figure*}[t]
    \centering
    \includegraphics[width=\linewidth]{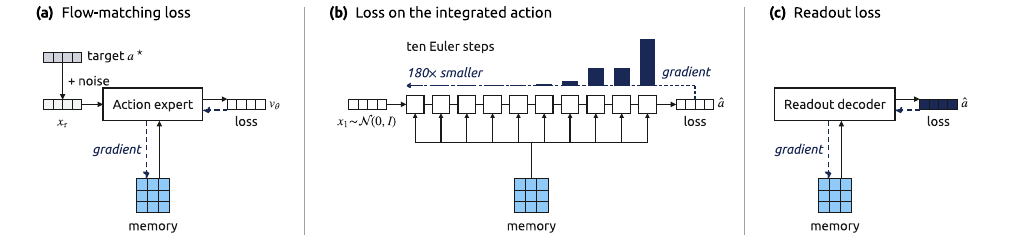}
    \caption{\textbf{Three losses for supervising the memory.}
    \textbf{(a)}~The flow-matching loss feeds the action expert a noised copy of the target action.
    \textbf{(b)}~A loss on the integrated action is backpropagated through ten Euler steps. Bars show the gradient norm at the state entering each step, relative to the last step.
    \textbf{(c)}~The readout loss supervises an action decoded from the memory in a single pass. Observation inputs are omitted.}
    \label{fig:rollout_loss}
\end{figure*}

To see the decision window of \cref{app:mif_theory} in numbers, we evaluate the two-mode case of Proposition~1 (\cref{fig:toyA}).
The demonstrated action $a\in\R^2$ lies near $(+3,0)$ or $(-3,0)$ with equal probability and a spread of 0.15, as when a robot reaches for one of two objects.
The observation is the same in both cases, and only the memory tells which of the two is demonstrated.
The half separation is $d=3$, which gives $\tau_s=0.75$ (\cref{eq:tau_s}).
No network is trained here, as both quantities in the figure follow from the distribution of the interpolant $x_\tau=(1-\tau)a+\tau\epsilon$.

The top row shows samples of $x_\tau$ for the two modes at five flow times.
The number above each card is the accuracy of the best guess of the mode from $x_\tau$ alone, which picks the nearer of the two cloud centers as in the proof of Proposition~1.
Near pure noise the two clouds overlap, and this guess is correct in 63\,\% of cases at $\tau=0.9$.
As $\tau$ decreases, the clouds separate, and the accuracy reaches 90\,\% at $\tau=0.7$ and exceeds 99\,\% at $\tau=0.5$.

The curve below is the memory gain $\mathcal G(\tau)$ of \cref{eq:mif_gain_proof}, computed from the exact posterior over the two modes and scaled to its maximum.
It peaks slightly below $\tau=1$ because of the weight $1/\tau^2$, and is large only where the clouds overlap.
Of the area under the curve, 76\,\% lies above $\tau_s$, which is a quarter of the flow-time axis, and less than 1\,\% lies below $\tau=0.5$.
At $\tau=0.5$, the gain has fallen to 1.7\,\% of its value at pure noise, within the bound of 9\,\% given by \cref{eq:vector_gain_bound}.
A training sample can thus reward the use of memory only when its flow time falls inside this window, and \cref{app:shortcut_rc5} locates the same window on RC5.

\section{Interpolant Shortcut on RC5}
\label{app:shortcut_rc5}

\paragraph{Estimating the memory gain.}
\Cref{fig:shortcut} estimates the memory gain of \cref{eq:decision_gain} on RC5 with the target slot as an oracle memory.
For every frame offset from the decision frame, we model the normalized action chunks of the demonstrations as a mixture of five Gaussians indexed by the target slot, with a pooled within-slot covariance.
We then compute the Bayes flow-matching loss from $x_\tau$ alone, once with the slot inferred and once with the slot given, and take the difference as the gain.
The largest eigenvalue of the decision-frame chunk covariance is $\Lambda=5.0$, which gives $\tau_s=0.69$ (\cref{app:mif_theory}).
At the decision frame, 94\,\% of the gain lies at $\tau\geq0.5$, and almost none lies below $\tau=0.3$.
Under the flow-time distribution of \PI, an estimated 8 to 13\,\% of training samples fall where the cue can lower the loss.
\Cref{app:present} extends this estimate with what the current observation reveals, which gives the map in \cref{fig:shortcut}.
The cards above the map embed the noised decision-frame chunks with t-SNE and report the cross-validated accuracy of naming the target cube from $x_\tau$ alone.
At high noise, this accuracy approaches 0.29, the share of the most common target slot, rather than the 0.20 of a uniform guess.

\paragraph{Reweighting the decision window.}
A natural alternative is to concentrate training on the decision window.
Starting from the trained \emph{memory-conditioned FM} policy, we continue training with the decision frames weighted twentyfold and their flow times restricted to $\tau\in[0.5,1)$ or $\tau\in[0.8,1)$.
The cue remains decodable from the memory (0.98 and 1.00), but the decision-frame chunk follows a swapped memory in only 0.21 and 0.22 of cases (chance 0.20).
More training signal within the window thus did not lead this policy to act on the cue.

\begin{table}[t]
    \centering\papertablefont
    \setlength{\tabcolsep}{6pt}
    \caption{\textbf{Supervising the memory on RC5.} Success rate (\%) over the number of flow steps for three policies trained on RC5 alone with a Gaussian source, which differ in the loss that supervises the memory (\cref{fig:rollout_loss}). Each entry uses 50 episodes, and chance is 20\,\%.}
    \label{tab:rollout_loss}
    \begin{tabularx}{\linewidth}{>{\raggedright\arraybackslash}Xcccc}
        \toprule
        Loss & 1 step & 2 steps & 5 steps & 10 steps \\
        \midrule
        Flow matching & 22 & 22 & 32 & 28 \\
        Integrated action & 0 & 0 & 0 & 28 \\
        Readout & \textbf{82} & \textbf{80} & \textbf{84} & \textbf{84} \\
        \bottomrule
    \end{tabularx}
\end{table}

\paragraph{Supervising integrated actions.}
A second alternative supervises the action obtained by integrating the flow from noise, which avoids the interpolant shortcut but requires backpropagation through every integration step (\cref{sec:method_ground}).
We compare three policies trained on RC5 alone with a Gaussian source, using the flow-matching loss, a loss on the action integrated over ten Euler steps, or the readout loss (\cref{fig:rollout_loss}).
Under the integrated-action loss, the gradient that reaches the first Euler step is about 180 times smaller than at the last step (\cref{fig:rollout_loss}b).
The resulting policy succeeds in 28\,\% of episodes with ten flow steps, as the flow-matching policy does (chance 20\,\%), and in none with fewer steps, whereas the readout loss reaches 84\,\% (\cref{tab:rollout_loss}).

\section{What the Current Observation Reveals}
\label{app:present}

\begin{figure*}[t]
    \centering
    \includegraphics[width=\linewidth]{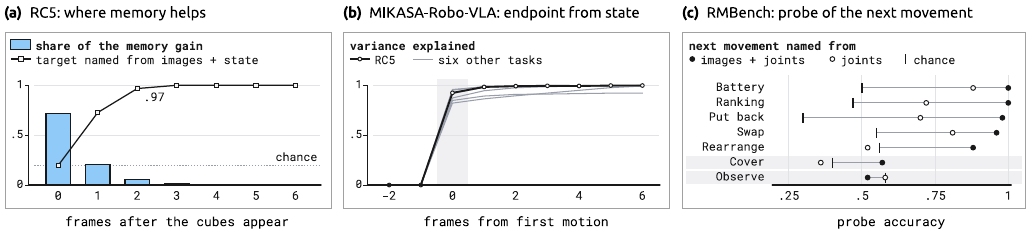}
    \caption{\textbf{What the current observation reveals.}
    \textbf{(a)}~RC5. Share of the estimated memory gain per frame after the cubes appear (bars), and accuracy of naming the target from images and robot state (line).
    \textbf{(b)}~MIKASA-Robo-VLA. Variance in the final hand position explained by the current robot state along demonstrations of seven tasks, with RC5 in black and the first moving frame in gray.
    \textbf{(c)}~RMBench. Leave-one-out accuracy of a nearest-centroid probe that names the next memory-dependent movement at its first frame, from images and joint state (filled) or joint state only (open), with chance marked by a tick.}
    \label{fig:present}
\end{figure*}

\providecolor{tabr1}{HTML}{C8E1FF}\providecolor{tabr2}{HTML}{E1F0FF}\providecolor{tabr3}{HTML}{F2F8FF}
\begin{table*}[t]
    \centering
    \papertablefont
    \fontsize{8}{9.6}\selectfont
    \setlength{\tabcolsep}{2.4pt}
    \caption{\textbf{Readout source versus Gaussian source on RMBench.} Success (\%) over 50 episodes per task for \ours and for the same policy trained and executed from a Gaussian source (Gauss.), with the same memory, readout supervision, and data. NFE is the number of flow steps, \emph{Readout} executes the readout $\hat a_t$ of \ours without the flow, and \emph{memory reset} clears the fast weights before every call at NFE\,=\,10. The \ours columns use the checkpoints of \cref{tab:main_rmbench}. \Cref{fig:mif}b plots the $\mathrm{M}(1)$ and $\mathrm{M}(n)$ means.}
    \label{tab:mif_nfe}
    \begin{tabular}{@{}lrrrrrrrrrrrrr@{}}
        \toprule
        & & \multicolumn{2}{c}{NFE\,=\,1} & \multicolumn{2}{c}{NFE\,=\,2} & \multicolumn{2}{c}{NFE\,=\,3} & \multicolumn{2}{c}{NFE\,=\,5} & \multicolumn{2}{c}{NFE\,=\,10} & \multicolumn{2}{c}{Memory reset} \\
        \cmidrule(lr){3-4}\cmidrule(lr){5-6}\cmidrule(lr){7-8}\cmidrule(lr){9-10}\cmidrule(lr){11-12}\cmidrule(lr){13-14}
        Task & Readout & \cellcolor{tabours}\ours & Gauss. & \cellcolor{tabours}\ours & Gauss. & \cellcolor{tabours}\ours & Gauss. & \cellcolor{tabours}\ours & Gauss. & \cellcolor{tabours}\ours & Gauss. & \cellcolor{tabours}\ours & Gauss. \\
        \midrule
        \multicolumn{14}{@{}l}{\emph{$\mathrm{M}(1)$ tasks}} \\
        Observe & 6 & \cellcolor{tabours}20 & 10 & \cellcolor{tabours}24 & 12 & \cellcolor{tabours}30 & 14 & \cellcolor{tabours}28 & 10 & \cellcolor{tabours}30 & 12 & \cellcolor{tabours}8 & 14 \\
        Put back & 28 & \cellcolor{tabours}22 & 82 & \cellcolor{tabours}66 & 92 & \cellcolor{tabours}84 & 94 & \cellcolor{tabours}100 & 82 & \cellcolor{tabours}100 & 92 & \cellcolor{tabours}0 & 0 \\
        Rearrange & 58 & \cellcolor{tabours}34 & 2 & \cellcolor{tabours}74 & 14 & \cellcolor{tabours}82 & 50 & \cellcolor{tabours}96 & 56 & \cellcolor{tabours}100 & 70 & \cellcolor{tabours}0 & 0 \\
        Swap & 80 & \cellcolor{tabours}90 & 2 & \cellcolor{tabours}94 & 2 & \cellcolor{tabours}98 & 10 & \cellcolor{tabours}92 & 10 & \cellcolor{tabours}96 & 16 & \cellcolor{tabours}2 & 16 \\
        Swap T & 58 & \cellcolor{tabours}78 & 66 & \cellcolor{tabours}78 & 58 & \cellcolor{tabours}72 & 66 & \cellcolor{tabours}80 & 66 & \cellcolor{tabours}82 & 62 & \cellcolor{tabours}0 & 2 \\
        Mean, $\mathrm{M}(1)$ & 46.0 & \cellcolor{tabours}\textbf{48.8} & 32.4 & \cellcolor{tabours}\textbf{67.2} & 35.6 & \cellcolor{tabours}\textbf{73.2} & 46.8 & \cellcolor{tabours}\textbf{79.2} & 44.8 & \cellcolor{tabours}\textbf{81.6} & 50.4 & \cellcolor{tabours}2.0 & 6.4 \\
        \midrule
        \multicolumn{14}{@{}l}{\emph{$\mathrm{M}(n)$ tasks}} \\
        Battery & 6 & \cellcolor{tabours}42 & 18 & \cellcolor{tabours}66 & 32 & \cellcolor{tabours}54 & 44 & \cellcolor{tabours}60 & 52 & \cellcolor{tabours}62 & 54 & \cellcolor{tabours}2 & 30 \\
        Ranking & 0 & \cellcolor{tabours}8 & 6 & \cellcolor{tabours}80 & 34 & \cellcolor{tabours}78 & 22 & \cellcolor{tabours}82 & 28 & \cellcolor{tabours}82 & 10 & \cellcolor{tabours}0 & 0 \\
        Cover & 32 & \cellcolor{tabours}32 & 0 & \cellcolor{tabours}48 & 0 & \cellcolor{tabours}48 & 0 & \cellcolor{tabours}72 & 0 & \cellcolor{tabours}84 & 0 & \cellcolor{tabours}0 & 0 \\
        Press & 10 & \cellcolor{tabours}14 & 0 & \cellcolor{tabours}52 & 0 & \cellcolor{tabours}54 & 0 & \cellcolor{tabours}58 & 0 & \cellcolor{tabours}70 & 2 & \cellcolor{tabours}0 & 2 \\
        Mean, $\mathrm{M}(n)$ & 12.0 & \cellcolor{tabours}\textbf{24.0} & 6.0 & \cellcolor{tabours}\textbf{61.5} & 16.5 & \cellcolor{tabours}\textbf{58.5} & 16.5 & \cellcolor{tabours}\textbf{68.0} & 20.0 & \cellcolor{tabours}\textbf{74.5} & 16.5 & \cellcolor{tabours}0.5 & 8.0 \\
        \midrule
        Mean, nine tasks & 30.9 & \cellcolor{tabours}\textbf{37.8} & 20.7 & \cellcolor{tabours}\textbf{64.7} & 27.1 & \cellcolor{tabours}\textbf{66.7} & 33.3 & \cellcolor{tabours}\textbf{74.2} & 33.8 & \cellcolor{tabours}\textbf{78.4} & 35.3 & \cellcolor{tabours}1.3 & 7.1 \\
        \bottomrule
    \end{tabular}
\end{table*}

\paragraph{RC5.}
The memory gain of \cref{app:memory_gain} vanishes once the current observation reveals the choice, and we measure how soon this happens.
A cross-validated logistic regression on the robot state and the top and wrist images gives a prior over the target slot at each frame, and we compute the Bayes velocity loss of the mixture model of \cref{app:shortcut_rc5} under this prior with and without the slot.
This probe identifies the selected cube at chance level (20\,\%) when the cubes appear, and with 97\,\% accuracy two frames later (\cref{fig:present}a).
As a result, the memory gain amounts to 10\,\% of the expected loss under the flow-time distribution of \PI, and would amount to 28\,\% if the observation revealed nothing.
Of the gain after the cubes appear, 97\,\% falls within the first three frames and 93\,\% at $\tau\geq0.5$.
\Cref{fig:shortcut} plots this estimate for the first seven frames after the cubes appear.

\paragraph{MIKASA-Robo-VLA.}
The same pattern appears on six other MIKASA-Robo-VLA tasks in which memory selects where to reach, namely SGP, ShellGameTouch, TIB, RememberColor3, RememberColor9, and RSC, where a cross-validated regression predicts the final end-effector position from the current robot state (\cref{fig:present}b).
From the first moving frame onward, the state explains at least 82\,\% of the variance in each of the seven tasks.

\paragraph{RMBench.}
The probe covers the seven RMBench tasks in which the memory-dependent step is a choice among a few movements.
On these tasks, we take the first frame of every movement whose instruction differs across the 50 demonstrations, and a leave-one-out nearest-centroid probe predicts that movement from the joint state alone or together with image features of the three cameras (\cref{fig:present}c).
With images and joint state, the probe identifies the next movement on Put back (0.98, chance 0.30), Swap (0.96, chance 0.55), Battery (1.00, chance 0.50), and Ranking (1.00, chance 0.47), and to a lesser extent on Rearrange (0.88, chance 0.56), whereas it stays near chance on Cover (0.57, chance 0.40) and Observe (0.52, chance 0.58).
On the former tasks, the demonstrated state thus already reveals the next movement once the robot's own earlier actions have left their trace in the scene.
A policy nevertheless needs memory to produce that trace in closed loop, as the memory resets in \cref{tab:mif_nfe} indicate.
Swap T and Press are not covered, as what must be remembered there is a placement pose and a number of presses rather than a choice of movement.
On Swap T, a nearest neighbor on the joint state predicts the placement pose with 0.16 of the error of predicting the mean.

\section{How Trained Policies Use Memory}
\label{app:policy_use}

This section details the memory-to-action analysis of \cref{fig:memrepr}, and reports the memory reset and flow-initialization results of \cref{sec:exp_main,sec:exp_ablation} per task.

\subsection{Frames, Policies, and Cue Decoding}
\label{app:memrepr_setup}

\paragraph{Frames and policies.}
\Cref{fig:memrepr} uses the decision frame of replayed RC5 demonstrations, defined as the first frame in which at least two cubes are visible in the top camera.
All reported numbers use the 78 frames in which all five cubes are in view, so that the chance of reaching the cued cube is one in five.
The two policies are \emph{memory-conditioned FM} (\cref{app:baselines}) and \ours, both trained on the same data from the same initialization.
Each frame is tested with its own memory and with the memories of four other episodes, one per other cue color, taken at the same step, which gives 78 own and 312 swapped memories.

\paragraph{Cue decoding.}
The linear probe of \cref{fig:memrepr}a,\,b standardizes the features, keeps the top 32 principal components, and fits a softmax regression.
All steps are fitted on the other episodes and, for a swapped memory, without both the receiving and the donor episode, so that the probe never sees the memory it classifies.
The projections shown in the figure are linear discriminant analyses for the cue, fitted on the same components and the same held-out split.
With 32, 64, or 100 components, the accuracy remains at least 0.97 for both policies.

\subsection{Action Choice and Controls}
\label{app:memrepr}

\paragraph{Choice rule.}
To decide which cube a generated action chunk heads to, we fit a linear map on the decision frames that predicts, from a cube's position in the top camera, the summed 8-step translation of a chunk that reaches it ($R^2=0.89$).
The chosen cube is the visible cube whose predicted translation is nearest to that of the chunk.
On the demonstrated chunks, this rule recovers the demonstrated target in 96\,\% of the 78 frames, also when the map is refit without the demonstration being judged.

\paragraph{Results.}
For each memory, we draw 16 source samples, from $\mathcal N(0,I)$ for \emph{memory-conditioned FM} and from $\hat a_t+0.1\epsilon$ for \ours.
\emph{Memory-conditioned FM} reaches the cued cube in 0.39 [0.33, 0.46] of actions with its own memory and in 0.16 [0.14, 0.17] with a swapped memory, whereas \ours reaches it in 0.96 [0.92, 1.00] and 0.95 [0.92, 0.98], respectively (\cref{fig:memrepr}c).
For \cref{fig:memrepr}d, replacing the noise changes the chosen cube in 0.56 [0.50, 0.60] of noise pairs for \emph{memory-conditioned FM} and in 0.01 [0.00, 0.03] for \ours, compared with 0.80 for two independent random choices among five cubes.
Replacing the memory changes the chosen cube in 0.11 of actions for \emph{memory-conditioned FM} and in 0.99 for \ours.
Brackets denote 95\,\% bootstrap intervals over frames.

\paragraph{Controls.}
The outcome does not depend on how a choice is scored, as scoring each memory by its most frequent choice instead of the share of actions gives 0.46 with the own memory and 0.14 with a swapped memory for \emph{memory-conditioned FM}, and 0.96 and 0.95 for \ours.
Unreadable memories or hesitant motions do not explain the gap either.
Restricting the analysis to memories from which the probe decodes the cue with probability at least 0.95 gives 0.41 [0.34, 0.48] and 0.16 [0.14, 0.18] for \emph{memory-conditioned FM}, and its chunks travel as far as the demonstrated chunks (median ratio 1.00).

\subsection{Memory Reset and Flow Initialization}
\label{app:source_interventions}
\label{app:regions}

\paragraph{Memory reset.}
Resetting the fast weights before every call lowers the success of \ours from 85.6\,\% to 33.2\,\% on MIKASA-Robo-VLA, with 32, 20, 34, 54, and 26\,\% on RC5, RSC, SGP, TIB, and IM, and from 78.4\,\% to 1.3\,\% on RMBench (\cref{tab:mif_nfe}).

\paragraph{Readout execution and flow refinement.}
Executing the readout $\hat a_t$ of \ours directly reaches 30.9\,\% mean success over the nine RMBench tasks, with 46.0\,\% on $\mathrm{M}(1)$ tasks and 12.0\,\% on $\mathrm{M}(n)$ tasks, whereas refining it through the flow reaches 78.4\,\% on the same checkpoints (\cref{tab:mif_nfe}).

\paragraph{Readout source and Gaussian source.}
\Cref{tab:mif_nfe} compares \ours with the same policy trained and executed from a Gaussian source, keeping the memory, readout supervision, and data.
At ten integration steps, \ours is ahead on all nine tasks, with 81.6\,\% against 50.4\,\% on $\mathrm{M}(1)$ tasks and 74.5\,\% against 16.5\,\% on $\mathrm{M}(n)$ tasks.
With fewer steps, \ours remains ahead on average at every number of steps, and Put back is the exception at one to three steps, where the Gaussian source succeeds more often.
With the memory reset before every call, the Gaussian-source policy retains 7.1\,\% mean success and \ours 1.3\,\%, which suggests that the successes of \ours rely on memory.

\end{document}